\documentclass[twoside]{article}

\usepackage[accepted]{aistats2026}
\usepackage[round]{natbib}

\usepackage{adjustbox}
\usepackage[toc,page,header]{appendix}
\newcommand{\D}{\mathcal{D}}
\newcommand{\Dtrain}{\mathcal{D}_{\mathrm{train}}}

\newcommand{\DL}{\mathcal{D}_{L,(f,\tau)}}
\newcommand{\DR}{\mathcal{D}_{R,(f,\tau)}}

\newcommand{\x}{\mathbf{x}}
\newcommand{\X}{\mathcal{X}}
\newcommand{\Y}{\mathcal{Y}}

\newcommand{\T}{\mathcal{T}}
\newcommand{\C}{\mathcal{C}}
\newcommand{\name}{\texttt{DEFT}\xspace}

\newcommand{\bmag}[1]{\textbf{\textcolor{blue!70}{#1}}}
\newcommand{\meanstd}[2]{\ensuremath{#1{\scriptstyle\,(\,#2\,)}}}

\usepackage{dsfont}

\usepackage{enumitem}
\usepackage{seqsplit}
\usepackage{amsmath}
\usepackage{amssymb}
\usepackage{mathtools}
\usepackage{amsthm}
\usepackage{lipsum}
\usepackage{soul}
\usepackage{listings}
\usepackage{graphicx}
\usepackage{microtype}
\usepackage{graphicx}
\usepackage{xspace}
\usepackage{xcolor}         
\usepackage{hyperref}
\usepackage[capitalize,noabbrev]{cleveref}
\crefname{appendix}{Appendix}{Appendices}
\Crefname{appendix}{Appendix}{Appendices}

\hypersetup{
    colorlinks=true,
    linkcolor=orange,
    filecolor=magenta,      
    urlcolor=cyan,
    citecolor=blue,
    pdftitle={eri},
    pdfpagemode=FullScreen,
    }
\usepackage{subcaption}
\usepackage{minitoc}

\usepackage{booktabs}

\usepackage{wrapfig}

\usepackage{multirow}

\usepackage{algorithm}
\usepackage{algorithmic}
\usepackage{bbm}

\usepackage{float}
\usepackage{tikz,forest}
\usetikzlibrary{arrows.meta}
\tikzset{
  treenode/.style = {shape=rectangle, rounded corners,
                     draw, align=center, 
                     minimum height=2ex, text depth=0.25ex,
                     top color=white, bottom color=blue!20},
  root/.style     = {treenode, font=\Large\rmfamily, bottom color=red!30},
  env/.style      = {treenode, font=\ttfamily\normalsize},
  leaf/.style     = {env, fill=orange!20, bottom color=orange!20, top color=orange!20}
}

\theoremstyle{plain}

\theoremstyle{definition}

\theoremstyle{remark}

\usepackage[framemethod=TikZ]{mdframed}
\mdfdefinestyle{contribution}{%
    middlelinecolor =   none,
    middlelinewidth =   1pt,
    backgroundcolor =   blue!10,
    roundcorner     =   5pt,
}

\mdfdefinestyle{feature}{%
    middlelinecolor=black,
    middlelinewidth=.8pt,
    backgroundcolor=blue!10,
    roundcorner=4pt,
    leftmargin=4pt,
    rightmargin=4pt,
    innerleftmargin=5pt,
    innerrightmargin=5pt,
    usetwoside=false
}

\mdfdefinestyle{takeaway}{%
    middlelinecolor =   none,
    middlelinewidth =   1pt,
    backgroundcolor =   teal!10,
    roundcorner     =   5pt,
}

\mdfdefinestyle{taxobox}{
  backgroundcolor=black!1,
  linecolor=black!55,
  linewidth=1pt,
  leftline=true, rightline=false, topline=false, bottomline=false, 
  roundcorner=2pt,
  innerleftmargin=3pt, innerrightmargin=3pt, innertopmargin=6pt, innerbottommargin=6pt,
  skipabove=8pt, skipbelow=10pt,
  needspace=3\baselineskip,
  frametitlebackgroundcolor=black!3,
  frametitleaboveskip=0.5ex, frametitlebelowskip=0.6ex,
  frametitlefont=\bfseries\scshape
}

\setlist[description]{
  style=nextline, leftmargin=0pt, labelsep=0.5em,
  font=\bfseries, itemsep=3pt, topsep=1pt
}

\usepackage[textsize=tiny]{todonotes}

\usepackage[utf8]{inputenc} 
\usepackage[T1]{fontenc}    
\usepackage{url}            
\usepackage{booktabs}       
\usepackage{amsfonts}       
\usepackage{nicefrac}       

\begin{document}

%

%

\twocolumn[

\aistatstitle{Interpretable DNA Sequence Classification via Dynamic Feature Generation in Decision Trees}

\runningauthor{Huynh, Kacprzyk, Sheridan, Bentley, van der Schaar}

\aistatsauthor{ Nicolas Huynh\textsuperscript{1}  \And Krzysztof Kacprzyk \And  Ryan Sheridan}
\aistatsaddress{ University of Cambridge \And University of Cambridge \And CU Anschutz Medical Campus} 
\aistatsauthor{David Bentley \And Mihaela van der Schaar}
\aistatsaddress{  CU Anschutz Medical Campus\And  University of Cambridge}]

\begin{abstract}
The analysis of DNA sequences has become critical in numerous fields, from evolutionary biology to understanding gene regulation and disease mechanisms. While deep neural networks can achieve remarkable predictive performance, they typically operate as black boxes. Contrasting these black boxes, axis-aligned decision trees offer a promising direction for interpretable DNA sequence analysis, yet they suffer from a fundamental limitation: considering individual raw features in isolation at each split limits their expressivity, which results in prohibitive tree depths that hinder both interpretability and generalization performance. We address this challenge by introducing \texttt{DEFT}, a novel framework that adaptively generates high-level sequence features during tree construction. \texttt{DEFT} leverages large language models to propose biologically-informed features tailored to the local sequence distributions at each node and to iteratively refine them with a reflection mechanism. Empirically, we demonstrate that \texttt{DEFT} discovers
human-interpretable and highly predictive sequence features across a diverse range of genomic tasks.
\end{abstract}

\vspace{15mm}

\vspace{-10pt}
\section{INTRODUCTION}
\textbf{DNA Sequence Analysis. }DNA sequences represent the fundamental code of life, storing the genetic instructions essential for the development and functioning of all organisms. 
In recent years, machine learning approaches have emerged as powerful tools for building predictive models with DNA sequences. Supervised learning methods have proven effective in various genomic tasks, from classifying promoter regions \citep{le2019classifying} and splice sites \citep{scalzitti2021spliceator, albaradei2020splice2deep} to predicting gene expression \citep{avsec2021effective}. However, predictive accuracy alone is \textit{not} sufficient for scientific utility, as \textit{transparency} is also important in biological research and clinical applications, where understanding the predictions is crucial for validation against known biological principles and discovery of biological features.

\textbf{Limitations of Black Boxes.} Despite their predictive performance, deep learning models  \citep{linder2025predicting, brixi2025genome} traditionally suffer from a lack of interpretability --- a limitation exacerbated by the scale of recent foundation models  \citep{dalla2025nucleotide}.  While post-hoc explanation methods, such as saliency maps \citep{simonyan2013deep} or visualization of attention maps \citep{avsec2021base} are frequently employed to probe these models, the faithfulness of post-hoc explanations to the model's predictions have been questioned, as different explanation methods can yield conflicting interpretations for the same prediction \citep{adebayo2018sanity, rudin2019stop}.

\textbf{Transparent Models.} Motivated by this observation, we ask the question: \textit{How do we design models that are inherently transparent, yet expressive for DNA sequence analysis?}  In the search for models that are transparent by design, we draw inspiration from common practices with tabular data, where decision trees are a \textit{ubiquitous} class of interpretable models \citep{murdoch2019definitions}. Crucially, they satisfy two interpretability properties that black-box models (even with post-hoc explanations) do not jointly possess \citep{lipton2018mythos}: \textit{simulatability} and \textit{decomposability}. 

\textbf{Axis-Aligned Trees.} Axis-aligned trees \citep{breiman1984classification, quinlan1986induction, quinlan2014c4} recursively partition the input space through binary decisions, based on the comparison between a specific feature and a learned threshold at every internal node. While they are widely used with tabular data (e.g. in finance or healthcare \citep{soleimanian2012application}), their standard formulation is \textit{flawed} in the context of DNA sequence analysis.

First, axis-aligned trees must grow deep to capture complex interactions between multiple sequence positions, as each split can only consider a single nucleotide position, thereby compromising the interpretability and generalization performance of the resulting tree.
   Second, manually crafting high-level variables offers an alternative but remains limited by existing biological knowledge and ignores \textit{local data characteristics} at each node during tree construction.

\textbf{Our Framework.} We address these limitations and answer our original question by introducing \name~ (Dynamic Engineering of Features in Trees), an \emph{interpretable} and \emph{expressive} tree-based model for DNA sequence classification.  At each internal node, \name~ automatically discovers \textit{high-level sequence features} that lead to discriminative splits, going beyond individual nucleotides. It is \textit{adaptive}, since feature generation is informed by the local data characteristics at each leaf of the tree. Furthermore, it incorporates \textit{domain knowledge} by favoring features that are biologically meaningful. \name~ achieves this by leveraging Large Language Models (LLMs) as adaptive feature generators, capitalizing on their in-context learning capabilities \citep{brown2020language} and domain knowledge acquired through extensive pretraining \citep{achiam2023gpt}. At each internal node, the LLM generates both an interpretable semantic representation and executable code for the proposed feature, guided by the partial tree structure and task-specific metadata. This generation process is embedded into an evolution-inspired optimization scheme in which the LLM iteratively refines candidate features through a \textit{self-reflection} mechanism based on the node splitting scores.
We evaluate \name~ across a broad range of tasks, showing that \name~ constructs decision trees with high-level, highly predictive features, outperforming other tree-based methods and rivaling black-box models while remaining \textit{transparent}.

\begin{mdframed}[style=contribution]
\bmag{Our contributions.} 
\bmag{(1)~Conceptually}, we propose \name, an interpretable model for DNA sequence classification that combines the transparency of decision trees with automated feature generation during tree construction.
\bmag{(2)~Technically}, we leverage Large Language Models (LLM) as adaptive feature generators that exploit local node context and task-specific metadata. We also employ an evolution-inspired optimization scheme in which the LLM iteratively refines candidate features through a reflection mechanism.
\bmag{(3)~Experimentally,} \name~ reveals \textit{interpretable} sequence features which are highly \textit{predictive} across diverse genomic tasks. 
 
\end{mdframed}

\section{RELATED WORKS}
\vspace{-10pt}
We summarize strands of research related to our work, with more details given in \Cref{app:extended_related_works}.

\textbf{Machine Learning for DNA Sequence Analysis.}
Machine learning methods have become essential tools for analyzing DNA sequences, with approaches ranging from classical models to deep neural networks. Traditional methods like position weight matrices and k-mer-based models \citep{stormo2000dna, ghandi2014enhanced} provide interpretable results but often lack expressivity. More recently, deep learning architectures, particularly convolutional and transformer networks \citep{linder2025predicting, brixi2025genome}, have demonstrated strong predictive performance across various genomic tasks such as transcription factor binding prediction \citep{alipanahi2015predicting, zeng2016convolutional, avsec2021base} and splice site prediction \citep{scalzitti2021spliceator, albaradei2020splice2deep, wang2019splicefinder}. However, these powerful models typically operate as black boxes, making it difficult to interpret their predictions. Although there exist various post-hoc interpretability methods \citep{simonyan2013deep}, they are not guaranteed to be faithful to the models they aim to explain \citep{rudin2019stop, adebayo2018sanity}.

\textbf{Decision Trees.} Decision trees are typically constructed greedily, finding at each node an optimal feature-threshold pair for splitting \citep{breiman1984classification, quinlan1986induction, quinlan2014c4}. These conventional methods are restricted to splits based on raw features, providing axis-aligned trees with limited expressivity. Exact optimization methods \citep{aglin2020learning, lin2020generalized}  also suffer from this bottleneck and can only be used in restricted settings. Multivariate decision trees \citep{murthy1994system, zhu2020scalable} allow splits to consider multiple features simultaneously (e.g. linear combinations of features for oblique trees). However, a common drawback of these methods is that they consider constrained feature spaces, are very challenging to optimize especially in high-dimensional settings, and assume continuous features.

\textbf{Applications of LLMs.} Recent works have leveraged LLMs for diverse tasks, including code evolution \citep{lehman2023evolution, brownlee2023enhancing}, optimization \citep{yang2024largelanguagemodelsoptimizers, liu2024evolution}, and feature engineering \citep{hanlarge, hollmann2024large, nam2024optimized}. However, \citep{hanlarge, hollmann2024large, nam2024optimized} have primarily focused on tabular datasets, while our work addresses DNA sequence analysis. Moreover, they do not account for the characteristics of the downstream model, whereas our approach integrates feature generation directly into tree induction, making it \textit{adaptive} to the local characteristics of nodes, which is a key contribution of \name. A recent line of works \citep{fallahpourbioreason} use LLMs to reason over biological modalities. The reasoning signal is a multi-step natural-language chain that takes a question in text and produces an answer, with explicit chain-of-thought (CoT). However, there is no guarantee that the CoT is faithful to the actual predictions. This contrasts \name, where the rationale is not a justification for a final prediction, but an internal CoT used to propose a feature at a given node; the final prediction is made by a decision tree that composes features.

\section{BACKGROUND} \label{sec:background}

We aim to build models that are transparent by design for DNA sequence analysis. To make this concrete, we require the following criteria \citep{lipton2018mythos}:

\begin{mdframed}[
  backgroundcolor=gray!10,  
  linecolor=gray!40,        
  linewidth=0.5pt,
  roundcorner=2pt,
  innertopmargin=6pt, innerbottommargin=6pt, innerleftmargin=8pt, innerrightmargin=8pt
]
\noindent\textbf{Simulatability.} A biologist should be able, in reasonable time, to carry out the computations needed to produce a prediction for a single example by following the model step by step.

\smallskip
\noindent\textbf{Decomposability.} Each part of the model (inputs/features, parameters, and intermediate computations) has a clear, human-meaningful semantics, and its contribution to the final prediction can be inspected in isolation (e.g. splits or rules correspond to specific sequence patterns).
\end{mdframed}
\vspace{-5pt}
Simulatability and decomposability are practically useful: they let the practitioner trace predictions to specific sequence features, check them against known biology, and plan targeted perturbations (e.g., CRISPR). While post-hoc methods can offer some insight on black boxes, they do \textit{not} satisfy both simulatability and decomposability. For example, feature attribution methods \citep{simonyan2013deep}, which are widely used with CNNs or Transformers, do not provide the explicit rules or computational steps the model used for its prediction. These considerations suggest using models that are interpretable by construction. Decision trees fit this category: their paths encode explicit rules, and their splits localize feature contributions.

\subsection{Top-down Tree Induction for Classification} Before describing our method, we recall the standard top-down tree induction procedure. Given a feature space $\X \subset \mathbb{R}^{d}$ and an output space $\Y$, decision tree induction is the process of learning a predictor $t: \X \rightarrow \Y$ from a dataset $\D_{\mathrm{train}} = \{ (\x_i, y_i) \}_ {i=1}^{N}$ of samples in $\X \times \Y$. The predictor $t$ is described by a tree structure which recursively partitions $\mathcal{X}$ into disjoint regions through feature-threshold splits at internal nodes, and each leaf region is associated with a prediction in $\Y$.

\textbf{Top-down Construction.} Decision trees are typically constructed top-down, with methods like \texttt{CART} \citep{breiman1984classification} and \texttt{ID3} \citep{quinlan1986induction} greedily building the tree one node at a time with axis-aligned splits. These methods select the optimal split at a given node by minimizing a score based on a node impurity measure $Q$ (e.g., Gini index, misclassification error, or information gain). 
Given a subset $\D$ of $\D_{\mathrm{train}}$, a feature map $f : \mathcal{X} \rightarrow \mathbb{R}$, and a threshold $\tau$, we denote by
\begin{align*}
    \DL &= \{(\x,y) \mid (\x,y) \in \D, f(\x) \leq \tau \} \\
    \DR &= \{(\x,y) \mid (\x,y) \in \D, f(\x) > \tau \}
\end{align*}
the subsets formed by splitting $\D$ with the feature map $f$ at threshold $\tau$.
 Furthermore, let $s$ be a scoring function based on the impurity measure $Q$, such that: \begin{equation}\label{eq:score} s(f,\tau,\D) = w^{l}_{f,\tau}Q(\DL) + w^{r}_{f,\tau}Q(\DR) \end{equation} where the weights $w^{l}_{f,\tau} = \frac{\vert \DL \vert}{\vert \D \vert}$ and $w^{r}_{f,\tau} = \frac{\vert \DR \vert}{\vert \D \vert}$ account for the relative sizes of the splits.

For any coordinate index $i \in [d]$, we denote by $\phi_{i}: \mathcal{X}\rightarrow \mathbb{R}$ the projection along the $i$-th feature. The induction process then finds the optimal pair of feature and threshold  $(i^*, \tau^*)$ for $\D$ by solving:
\begin{equation} \label{eq:splitting_condition}
    (i^*, \tau^*) \in \arg\min_{(i,\tau) \in [d]\times\mathbb{R}} s(\phi_{i}, \tau, \D)
\end{equation}

Equipped with this optimization procedure, top-down approaches progressively grow the tree, starting from a single root node containing the training set $\Dtrain$. Given a leaf $v$ in the partially constructed tree corresponding to the subset $\D \subset \Dtrain$ which satisfies all splitting conditions from root to $v$, top-down approaches find an optimal split $(i^*, \tau^*)$ for $\D$ using the criterion defined in \cref{eq:splitting_condition}. The leaf $v$ then spawns two child nodes containing $\mathcal{D}_{L,(\phi_{i^*},\tau^{*})}$ and $\mathcal{D}_{R,(\phi_{i^*},\tau^{*})}$ respectively, and the induction process continues with the updated tree, until a stopping criterion is met (e.g. when a maximum depth is reached).

\subsection{Limitations of Axis-Aligned Trees for DNA Sequence Analysis}
The transparency of decision trees with a limited depth \citep{murdoch2019definitions}, coupled with their predictive performance, explains their widespread use with \textit{tabular data}. However, naively leveraging traditional axis-aligned decision trees for \textit{DNA sequence analysis} presents several key challenges, which we now detail.

\textbf{Limited Expressivity of Raw Sequence Features.} 
Decision trees can be trained on raw DNA sequence features after basic preprocessing, such as one-hot encoding of nucleotides or treating them as ordinal variables \citep{hastie2009elements}. However, \cref{eq:splitting_condition} reveals a fundamental limitation: the induction process considers \textit{positions in isolation} at each node, unable to capture interactions between different sequence positions simultaneously. This limitation creates a tension between \textit{expressivity} and \textit{interpretability}. Since each split considers only a single position, trees must grow deep to capture complex sequence patterns that depend on multiple nucleotide positions. This compromises the interpretability that makes decision trees appealing in the first place, since predictions are described by long sequences of splitting conditions. Furthermore, this complexity also impacts generalization performance, as deeper trees might capture spurious patterns which do not generalize well at test time. 

\begin{figure} 
    \centering
    \includegraphics[width=0.3\textwidth]{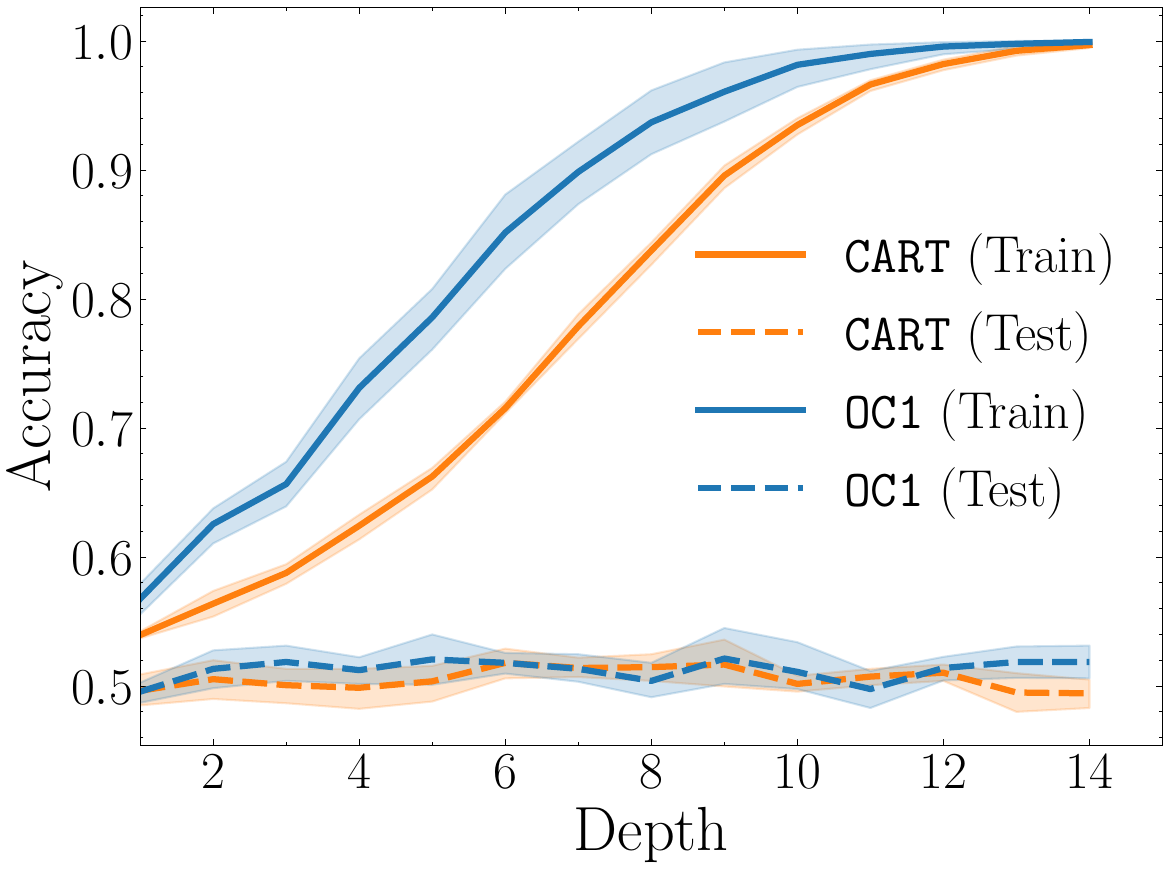} 
    \caption{\textbf{Limitations of conventional trees.} Training and test accuracies versus tree depth for motif detection, mean and confidence intervals at $95\%$ reported for $5$ seeds. Conventional decision trees must grow deep to achieve a high training accuracy, and they do not learn patterns which generalize at test time.}
    \label{fig:motif_detection_example}
\end{figure}
\underline{\textit{An example.}} We illustrate these limitations with a simple example. Given a dataset of DNA sequences, we consider the task of predicting whether or not the motif "TATA" is present, which requires considering multiple sequence positions simultaneously. As we show in \cref{fig:motif_detection_example}, \texttt{CART} struggles with this seemingly simple task. While it can achieve high training accuracy by growing deep enough to memorize specific position-by-position patterns, the test accuracy remains poor. This is a critical limitation since many biologically meaningful patterns emerge from the joint consideration of multiple nucleotides positions, with examples including secondary structure formation or binding site motifs. While multivariate trees \citep{murthy1994system, zhu2020scalable} offer improved expressivity by splitting on linear combinations of raw features, the space of features they can explore is limited, potentially missing out on more complex combinations that could lead to better generalization. We show this in \cref{fig:motif_detection_example} with \texttt{OC1} (oblique decision tree induction \citep{murthy1994system}) achieving better training performance than \texttt{CART}, but it does not find patterns which generalize at test time.

\textbf{Constraints of Manual Feature Engineering.}
 A potential solution to the aforementioned limitation is to manually craft higher-level features that capture known biological patterns and sequence motifs. However, this approach faces two fundamental limitations. First, it remains inherently constrained by current \textit{biological knowledge}, potentially missing important but undiscovered sequence patterns—a significant drawback when studying problems whose underlying biological mechanisms are not yet understood. Second, engineered features, such as $k$-mers features \citep{ghandi2014enhanced}, are typically designed in a model-agnostic way, lacking \textit{adaptivity} to local data characteristics that emerge during tree induction, and which should inform feature generation.

\definecolor{myorange}{HTML}{EA7203}
\definecolor{myblue}{HTML}{3C92D2}

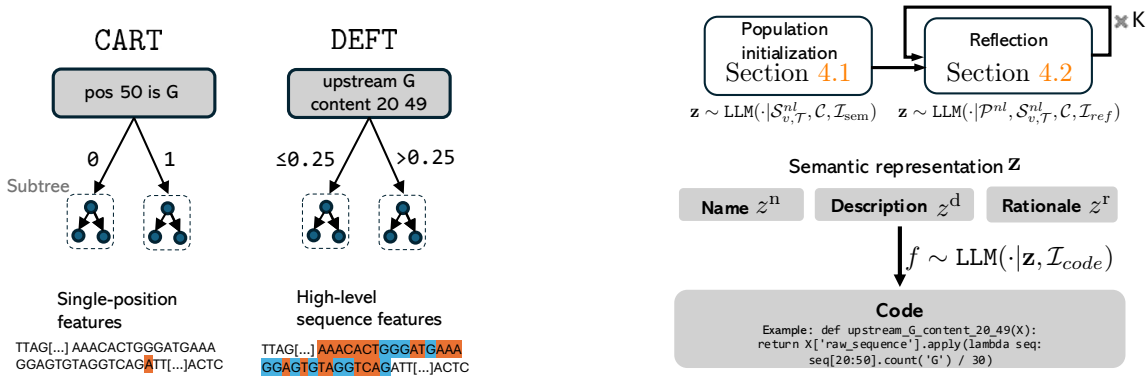
\begin{figure*}
    \centering
    \begin{subfigure}[b]{0.5\textwidth}
        \centering
        \includegraphics[height=5.5cm]{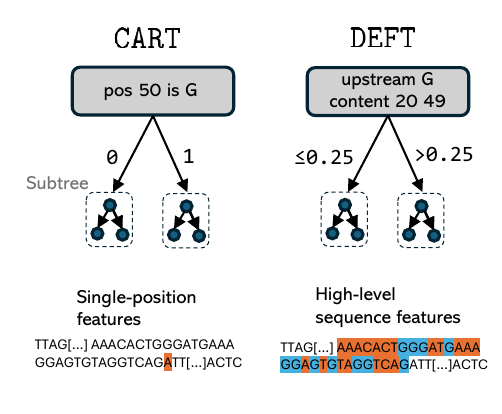}
        \label{fig:subfig-a}
    \end{subfigure}
    \hfill
    \begin{subfigure}[b]{0.45\textwidth}
        \centering
        \begin{tikzpicture}
            \node[anchor=south west,inner sep=0] (image) at (0,0) 
                {\includegraphics[height=5.5cm]{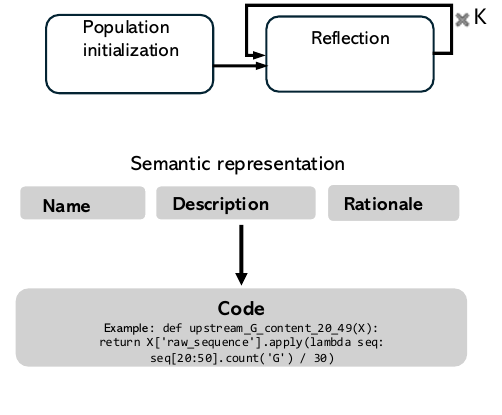}};
            
            \begin{scope}[x={(image.south east)},y={(image.north west)}]
                \node at (0.26,0.83) {\cref{subsec:init}};

                \node at (0.25, 0.73)[scale=0.7]
                {$\mathbf{z} \sim \texttt{LLM}(\cdot \vert \mathcal{S}_{v,\mathcal{T}}^{nl}, \mathcal{C}, \mathcal{I}_{\mathrm{sem}})$};

                 \node at (0.7, 0.73)[scale=0.7]
                {$\mathbf{z} \sim \texttt{LLM}(\cdot \vert \mathcal{P}^{nl}, \mathcal{S}_{v,\T}^{nl}, \mathcal{C}, \mathcal{I}_{ref})$};
                
                \node at (0.7, 0.83)
                {\cref{subsec:ref}};

                \node at (0.71, 0.607)
                {$\mathbf{z}$};

                \node at (0.22, 0.51)
                {$z^{\text{n}}$};

                \node at (0.58, 0.51)
                {$z^{\text{d}}$};

                 \node at (0.88, 0.51)
                {$z^{\text{r}}$};

                \node at (0.7, 0.38)
                {$f \sim \texttt{LLM}(\cdot \vert \mathbf{z}, \mathcal{I}_{code})$};
               
            \end{scope}
        \end{tikzpicture}
        \label{fig:subfig-b}
    \end{subfigure}
    \caption{\name~ is a tree-based method for interpretable DNA sequence analysis. \textbf{Left}: \name~ discovers high level sequence features that can consider multiple positions simultaneously, contrasting \texttt{CART}. For example, the feature \texttt{upstream\_G\_content\_20\_49} operates on a window of positions highlighted in \textcolor{myorange}{orange} (non-G nucleotides) and \textcolor{myblue}{blue} (G nucleotides). \textbf{Right}: It leverages LLMs to generate candidate features at each node in a two-step process: first outputting semantic representations, then converting them into Python executable code. \name~ conditions the generation of features on the partial tree structure and task-specific metadata.}
    \label{fig:both-subfigs}
\end{figure*}

\section{METHOD}
 The limitations of conventional decision trees detailed in \cref{sec:background} —namely their restricted expressivity when using raw features, and the constraints of manual feature engineering—show the need for a novel framework for interpretable DNA sequence analysis. To address these challenges directly, we introduce \texttt{DEFT}, which combines the \textit{transparency} of tree structures while ensuring \textit{expressivity} through automatically generated sequence features which yield discriminative splits during tree construction. 

\textbf{Adaptive Feature Generation with LLMs.} Searching over the space $\mathbb{R}^{\mathcal{X}}$of possible feature maps at every leaf node during tree induction is highly non-trivial, due to its combinatorial nature and the high dimensionality of DNA sequences. Instead, our key insight is to leverage Large Language Models (LLMs) as adaptive feature generators, capitalizing on their prior knowledge acquired through pretraining and their in-context learning capabilities. Furthermore, they are powerful hypothesis generators \citep{wang2023hypothesis}, making them particularly suitable for exploring the combinatorial space of feature maps.

\textbf{Method Overview.} Our method \name~ performs tree induction in a top-down manner, progressively growing the tree starting from the root node. At every leaf node of the partially constructed tree, \name~ generates an initial set of candidate feature maps through a two-step process, first producing semantic descriptions and then generating corresponding executable code for feature computation. \name~ then employs an evolution-inspired optimization scheme where the LLM iteratively refines the candidate features through self-reflection to optimize the splitting criterion. In what follows, we detail each of these components in turn.

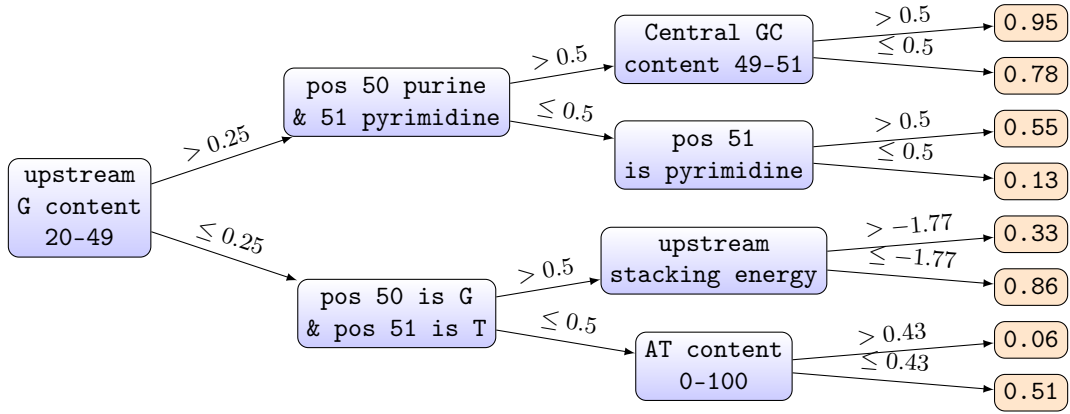
\begin{figure*}[!t]
    \centering
    \begin{adjustbox}{max width=0.88\linewidth}
\begin{tikzpicture}[
grow                     = right,
level 1/.style={sibling distance=8em},
level 2/.style={sibling distance=4em},
level 3/.style={sibling distance=2em},
level distance           = 12em,
edge from parent/.style  = {draw, -latex},
every node/.append style = {font=\footnotesize},
sloped
]

\node [env] {upstream\\ G content \\ 20-49}
    child { node [env] {pos 50 is G \\ \& pos 51 is T}
        child { node [env] {AT content\\ 0-100}
            child { node [leaf] {0.51}
                edge from parent node [above] {$\leq 0.43$}}
            child { node [leaf] {0.06}
                edge from parent node [above] {$> 0.43$}}
            edge from parent node [above] {$\leq 0.5$}}
        child { node [env] {upstream \\ stacking energy}
            child { node [leaf] {0.86}
                edge from parent node [above] {$\leq -1.77$}}
            child { node [leaf] {0.33}
                edge from parent node [above] {$> -1.77$}}
            edge from parent node [above] {$> 0.5$}}
        edge from parent node [above] {$\leq 0.25$}}
    child { node [env] {pos 50 purine \\ \& 51 pyrimidine}
        child { node [env] {pos 51 \\ is pyrimidine}
            child { node [leaf] {0.13}
                edge from parent node [above] {$\leq 0.5$}}
            child { node [leaf] {0.55}
                edge from parent node [above] {$> 0.5$}}
            edge from parent node [above] {$\leq 0.5$}}
        child { node [env] {Central GC \\ content 49-51}
            child { node [leaf] {0.78}
                edge from parent node [above] {$\leq 0.5$}}
            child { node [leaf] {0.95}
                edge from parent node [above] {$> 0.5$}}
            edge from parent node [above] {$> 0.5$}}
        edge from parent node [above] {$> 0.25$}}
;

\path[use as bounding box]
  ([xshift=-2mm,yshift=-2mm]current bounding box.south west)
  rectangle
  ([xshift= 2mm,yshift= 2mm]current bounding box.north east);

\end{tikzpicture}
\end{adjustbox}
    \caption{\textbf{Decision tree constructed by $\name$ for the Pol II dataset.} $\name$ discovers high-level sequence \textcolor{blue}{features}. We also report the \textcolor{orange}{leaves' predictions}. Values along the edges correspond to the threshold at each split.}
    \label{fig:decision_tree}
\end{figure*}

\subsection{Initializing the Population of Candidate Features} \label{subsec:init}
 Given a partially constructed tree $\T$ and a leaf node $v$ associated with the subset $\D \subset \Dtrain$, we seek a feature map $f^*: \X \rightarrow \mathbb{R}$ that is discriminative for $\D$. Unlike traditional top-down approaches that only search over the set $\{\phi_{i}\}_{i=1}^{d}$ of raw feature maps (cf. \cref{eq:splitting_condition}), $\name$~ explores a richer space of feature maps in $\mathbb{R}^{\mathcal{X}}$ by considering the unique characteristics of $\D$.
 It generates initial candidate features through a two-step process, first generating semantic representations describing the candidate features, and then obtaining executable code to compute these candidate features.

\textbf{Step 1. Obtaining Semantic Representations.} \name~ first generates $M$ semantic representations $\{ \mathbf{z}_{j} \}_{j=1}^{M}$, where each triplet $\mathbf{z}_j = (z^{\text{r}}_j, z^{\text{n}}_j, z^{\text{d}}_j, )$ provides a human-interpretable specification of a candidate feature: $z^{\text{r}}_j$ provides a rationale to ground the feature in potential biological relevance or observed data patterns (e.g., \emph{"GC-rich regions often indicate regulatory elements"}), $z^{\text{n}}_j$ provides a concise name (e.g., \emph{"GC content"}), and $z^{\text{d}}_j$ provides a precise description for unambiguous implementation (e.g., \emph{"percentage of G and C nucleotides in the first 20 positions"}). The goal of the rationale $z^{\text{r}}_j$ is to anchor the LLM's feature proposal by requiring an explicit justification, encouraging the generation of features that are not only syntactically valid but also plausible within the given biological context. To generate semantic representations suited to the characteristics of $\D$, we incorporate information from the partial tree structure $\T$ and the leaf $v$.  Specifically, we represent the path from the root node to the leaf $v$ in $\T$ as a sequence of splitting conditions  $\mathcal{S}_{v, \T}$  defined by $\mathcal{S}_{v, \T} = \{ (f_{l}, \tilde{\mathbf{z}}_{l}, o_{l}, \tau_{l}) \}_{l=1}^{L}$ where $L$ is the path length, and each tuple consists of a feature map $f_{l}: \X \rightarrow \mathbb{R}$, its semantic representation $\tilde{\mathbf{z}}_{l}$, a comparison operator $o_{l} \in \{ \leq, > \}$, and a threshold $\tau_{l}$. We then serialize this node context in natural language to obtain a prompt $\mathcal{S}_{v,\T}^{nl}$, using the few-shot template: "\{$\tilde{z}_l^{n}$\}~ \{$o_{l}$\}~\{$\tau_{l} $\}~ (\{$\tilde{z}_l^{d}$\})".
In addition to the node context, we also guide the generation with a task context $\C$ in natural language that describes the input space $\X$, output space $\Y$, and prediction task.
We then sample semantic representations of candidate features from the LLM as  $\mathbf{z}_{j} \sim \texttt{LLM}(\cdot \vert \mathcal{S}_{v,\T}^{nl}, \C, \mathcal{I}_{\mathrm{sem}})$, where $\mathcal{I}_{\mathrm{sem}}$ contains the generation instructions that specify the expected format.

\textbf{Step 2. Obtaining Executable Code.} For  each semantic representation $\mathbf{z}_{j}$, we generate an executable implementation of the corresponding feature by prompting the LLM to translate the natural language representation into Python code as $f_j \sim \texttt{LLM}(\cdot \vert \mathbf{z}_{j}, \mathcal{I}_{code})$, where $\mathcal{I}_{code}$ contains instructions for producing valid Python code that computes the feature value for any input in $\X$. 

We then define the initial population of candidate features $P = \{ ( f_j, \mathbf{z}_{j} ) \}_{j=1}^K \cup \{ (\phi_{i}, \bar{\mathbf{z}}_{i}) \}_{i=1}^{d}$, where we incorporate the raw features and their associated semantic representations in addition to the LLM-generated features.

\subsection{Iterative Improvement with Reflection} \label{subsec:ref}
While the contextual information provided by $S_{v, \T}$ and $\mathcal{C}$ guides the generation of the initial candidate features, it may not be sufficient to obtain highly discriminative features. We therefore propose an evolution-inspired optimization scheme that iteratively improves feature quality through LLM-based reflection.

Given a population $P$ of features and their semantic representations for a node with dataset $\D$, we first evaluate the discriminative power of the features by computing for each $(f, \mathbf{z}) \in P$ the score $\eta = \min_{\tau \in \mathbb{R}} s(f,\tau,D)$, where $s$ depends on an impurity measure $Q$ (cf. \cref{eq:score}). We then create a few-shot prompt $\mathcal{P}^{nl}$ by serializing each feature $(f, \mathbf{z})$ with its score $\eta$ following the template \emph{"Score: $\{\eta\}$, Feature name: $\{z^{\text{n}} \}$, Feature description: $\{ z^{\text{d}} \}$, Feature code: $\{f\}$"}.

To obtain better features, we define a set of instructions $\mathcal{I}$ comprising two distinct prompt instructions: one for \textit{exploration}, which directs the LLM to propose features distinct from $P$, and one for \textit{exploitation}, which guides the LLM to analyze and refine patterns from the highest-performing in-context features. These instructions also incorporate constraints regarding the interpretability of the generated features, an aspect that we investigate in \cref{subsec:tradeoff}. For each instruction $\mathcal{I}_{ref} \in \mathcal{I}$, we generate a set of $M$ semantic representations as  $\mathbf{z}'_{m} \sim \texttt{LLM}(\cdot \vert \mathcal{P}^{nl}, \mathcal{S}_{v,\T}^{nl}, \mathcal{C}, \mathcal{I}_{ref})$. Each $\mathbf{z}'_{m}$ is then transformed into executable code, yielding a population $P'$. The $M$ solutions from $P \cup P'$ with lowest scores are selected to form the next population, and this optimization process repeats for $K$ iterations. Finally, we select the feature that achieves the minimum score in the final population as the splitting feature for the current node $v$. Since we include the raw features $\{ \phi_{i} \}_{i=1}^{d}$ in the initial population, we have the guarantee that the feature selected by \name~ at the end of the optimization procedure will be at least as discriminative as the feature that an axis-aligned tree would select. 

We summarize the different steps for feature generation in \cref{alg:feature_generation} in the Appendix, and provide details on the prompts in \Cref{app:experimental_details}. Once the tree is fully constructed, predictions for new samples are obtained by traversing the tree from the root to a leaf node, as detailed in \cref{alg:prediction}, and comparing at each node the feature value against the node's learned threshold, analogous to conventional decision tree inference.

\textbf{Computation Cost.} A discussion of the computation cost can be found in \Cref{app:computational_overhead}. In brief, the cost scales linearly with the number of reflection steps $K$. Furthermore, since feature generation at the same tree depth is independent, the process is highly parallelizable.

\section{EXPERIMENTS} \label{sec:experiments}

\begin{figure*}[h]
    \centering

    \begin{subfigure}[t]{0.32\linewidth}
        \centering
        \includegraphics[width=\linewidth]{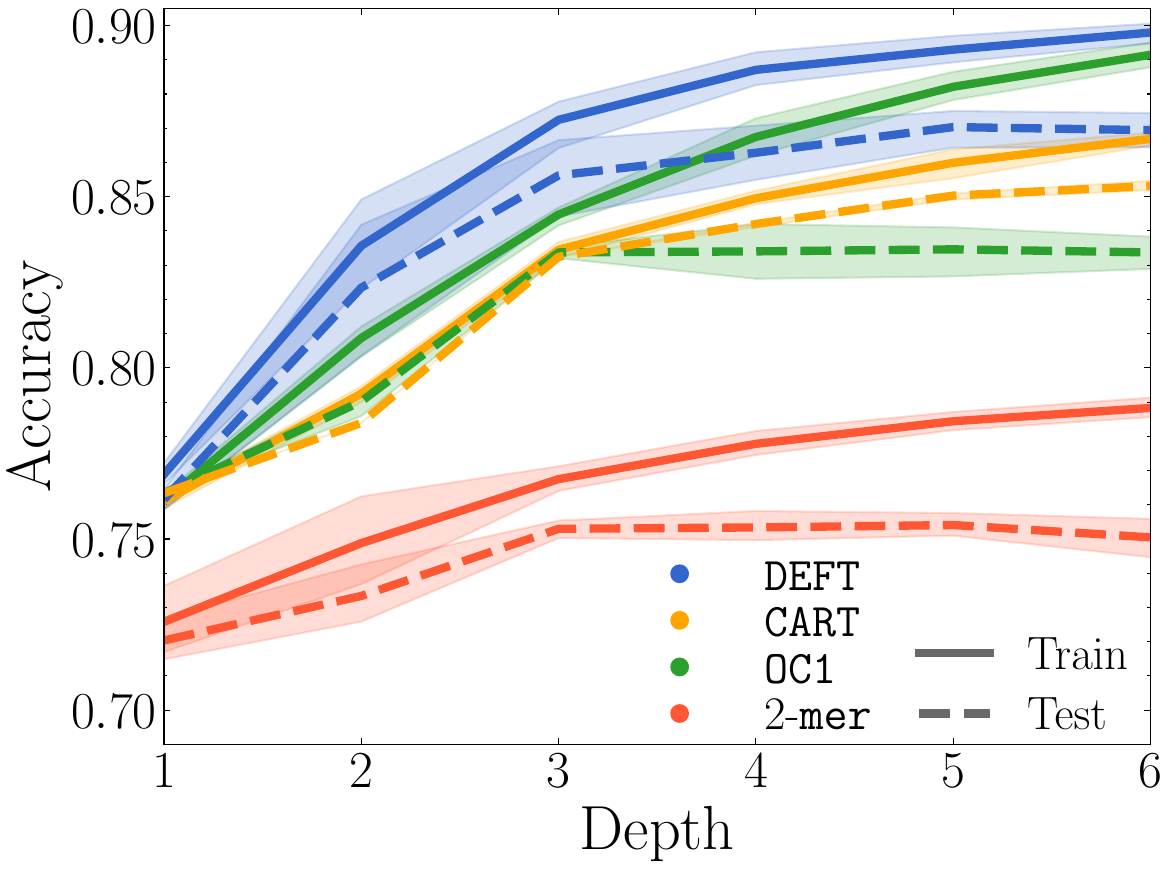}
        \caption{Pol II Pausing}
        \label{fig:comparison-with-baselines:polymerase}
    \end{subfigure}\hfill
    \begin{subfigure}[t]{0.32\linewidth}
        \centering
        \includegraphics[width=\linewidth]{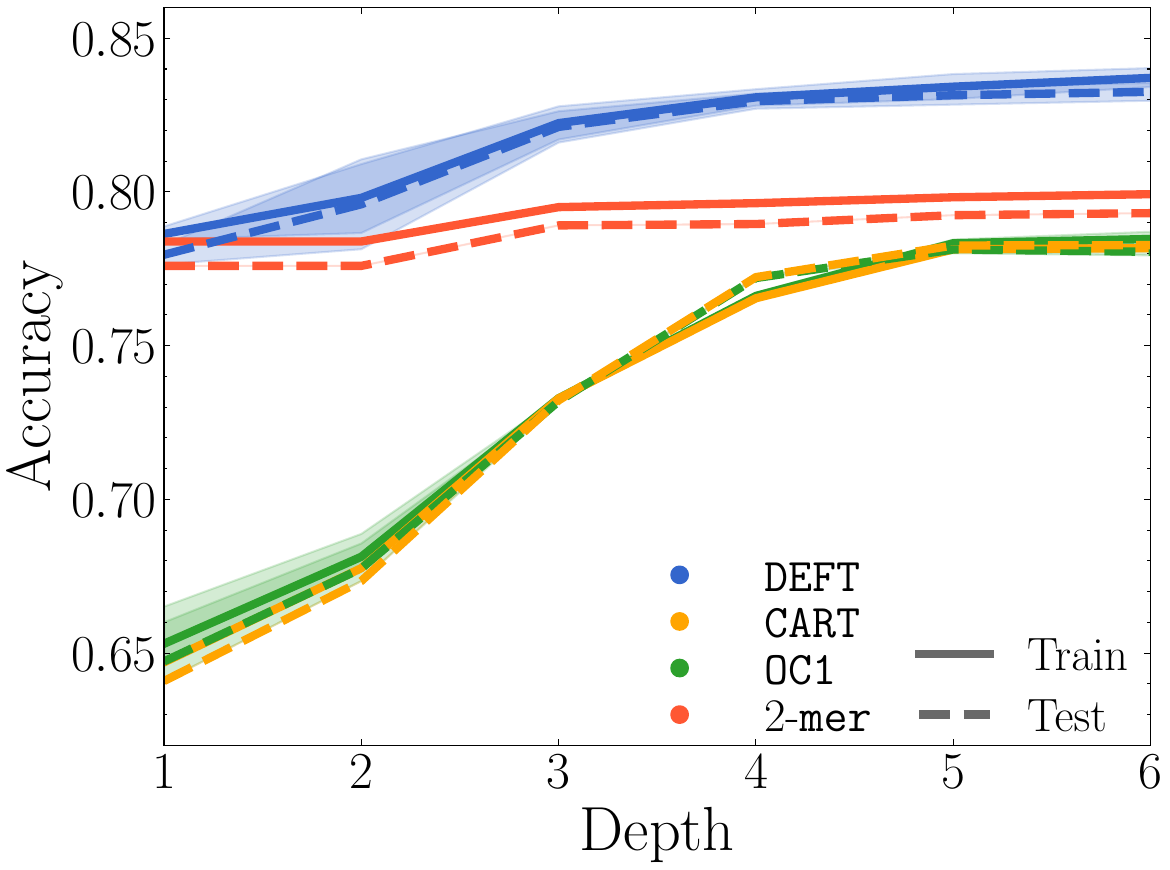}
        \caption{Promoters}
        \label{fig:comparison-with-baselines:promoters}
    \end{subfigure}\hfill
    \begin{subfigure}[t]{0.32\linewidth}
        \centering
        \includegraphics[width=\linewidth]{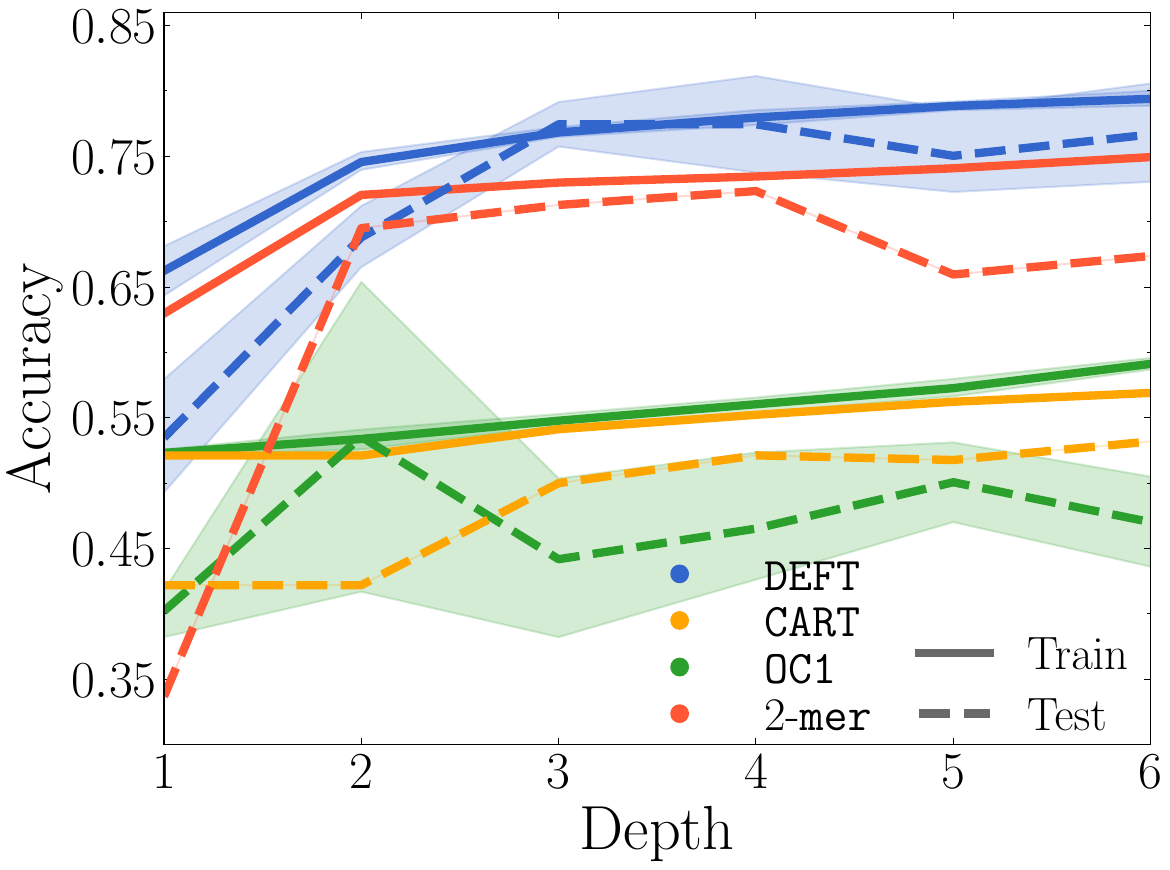}
        \caption{Enhancers}
        \label{fig:comparison-with-baselines:enhancers}
    \end{subfigure}

    \caption{\textbf{Performance comparison against tree-based baselines.} Training and test accuracies across varying depths
    (mean $\pm$ 95\% CIs over 5 seeds). \name consistently outperforms the tree-based baselines.}
    \label{fig:comparison_with_baselines_accuracy}
\end{figure*}

In this section, we evaluate \name~across diverse genomics tasks. In \cref{subsec:exp:performance}, we benchmark predictive performance against tree-based baselines and black-box models, showing that \name~ generates highly predictive sequence features. In \cref{subsec:discussion_features}, we analyze the features learned by \name, introducing a taxonomy that reveals how they adapt to dataset-specific characteristics. In \cref{subsec:tradeoff}, we show that \name~ enables control of the balance between interpretability and discriminative power of the discovered features. In \cref{subsec:exp:ablations}, we perform ablation studies that provide insight into the key mechanisms contributing to \name~'s performance. Code to reproduce our experiments can be found at \url{https://github.com/nicolashuynh/deft}.

\textbf{Datasets.} We evaluate \name on a diverse set of genomics tasks: \textit{Pol II Pausing} \citep{mayer2017pause}, predicting RNA polymerase II pausing from DNA sequence context; \textit{Promoters} \citep{grevsova2023genomic}, identifying human non-TATA promoters; and \textit{Enhancers} \citep{ernst2016genome}, classifying enhancer activity from massively parallel reporter assay constructs. More details on the datasets can be found in \Cref{app:experimental_details}.

\textbf{Experimental details.} In what follows, we use \texttt{gpt-4o} version \texttt{1001} as the underlying LLM. We use a population size $M=10$ and $K=20$ reflection steps per node. The Gini index serves as the splitting criterion at each node and we set a minimum number of samples per leaf equal to $1\%$ of the size of the training set to prevent overfitting.  \cref{app:experimental_details} provides more details on the hyperparameters and experimental setting, and code can be found in the supplementary material.

\vspace{-2mm}
\subsection{Predictive Performance of \name} \label{subsec:exp:performance}
\vspace{-2mm}

\textbf{Methodology.} We first compare \name  against interpretable tree-based baselines:
1) \textit{CART} \citep{breiman1984classification}, standard axis-aligned decision tree induction (with one-hot encoded nucleotides at each sequence position \footnote{Treating nucleotides as ordinal variables empirically yielded inferior results to one-hot encoding.}) ;
2)~ \textit{OC1} \citep{murthy1994system}, oblique decision trees;
and 3) \textit{CART} with $2$-mer features, where counts of all possible dinucleotides serve as input features. All methods are evaluated across varying maximum tree depths $d \leq 6$
, with a minimum of $1\%$ of training samples per leaf to mitigate overfitting. We also consider a logistic regression baseline using $2$-mers features.
While the focus of this work is on developing \textit{transparent} models for DNA sequence analysis, we contextualize \name~'s performance by comparing it with non-transparent models: deep-learning models (\textit{Convolutional Neural Networks}, \textit{Transformers}), and ensembles ( \textit{XGBoost} with $2$-mers).

\begin{table*}[t]
\centering
\small
\setlength{\tabcolsep}{6pt}
\caption{\textbf{Performance comparison against black boxes and ensemble baselines.}}
\label{tab:blackbox-comparison}
\begin{tabular}{lcccccc}
\toprule
& \multicolumn{2}{c}{Pol II Pausing} & \multicolumn{2}{c}{Promoters} & \multicolumn{2}{c}{Enhancers} \\
\cmidrule(lr){2-3}\cmidrule(lr){4-5}\cmidrule(lr){6-7}
Method & Acc. & AUPRC & Acc. & AUPRC & Acc. & AUPRC \\
\midrule
\texttt{CNN} & \meanstd{0.868}{0.007} & \meanstd{0.933}{0.006} & \meanstd{0.828}{0.005} & \meanstd{0.931}{0.002} & \meanstd{0.790}{0.065} & \meanstd{0.964}{0.006} \\
\texttt{Transformer} & \meanstd{0.876}{0.016} & \meanstd{0.942}{0.002} & \meanstd{0.835}{0.006} & \meanstd{0.933}{0.005} & \meanstd{0.704}{0.045} & \meanstd{0.952}{0.007} \\
\texttt{XGBoost}  & \meanstd{0.759}{0.005} & \meanstd{0.848}{0.002} & \meanstd{0.822}{0.000} & \meanstd{0.925}{0.000} & \meanstd{0.777}{0.000} & \meanstd{0.965}{0.000} \\
\midrule 
\texttt{Log.} \texttt{Reg.} & \meanstd{0.769}{0.002} & \meanstd{0.857}{0.001} & \meanstd{0.796}{0.000} & \meanstd{0.901}{0.000} & \meanstd{0.730}{0.000} & \meanstd{0.945}{0.000} \\
\texttt{CART} & \meanstd{0.750}{0.007} & \meanstd{0.830}{0.003} & \meanstd{0.793}{0.000} & \meanstd{0.897}{0.000} & \meanstd{0.674}{0.000} & \meanstd{0.939}{0.000} \\
\name & \meanstd{0.869}{0.007} & \meanstd{0.912}{0.006} & \meanstd{0.833}{0.004} & \meanstd{0.926}{0.004} & \meanstd{0.767}{0.049} & \meanstd{0.948}{0.012} \\
\bottomrule
\end{tabular}

\vspace{-2pt}
\raggedright \footnotesize
\end{table*}

\textbf{Results.} We compare in \cref{fig:comparison_with_baselines_accuracy} the training and test accuracies of \name~ and tree-based baselines across varying depths for $5$ different seeds. \name~ consistently achieves superior performance on both training and test sets for the different depths. In \Cref{tab:blackbox-comparison}, we compare \name against the non-transparent baselines, reporting test results (mean and std for 5 seeds) and showing that \name is competitive while remaining interpretable by design, closing the gap between transparent and black-box models. We also report additional results in \Cref{app:additional_results}, e.g. other performance metrics, a comparative analysis with the Nucleotide Transformer \citep{dalla2025nucleotide}, results with an open-source LLM (\texttt{gpt-oss}), and performance gains obtained when ensembling \name's trees.  We also compare \name with dataset-specific deep-learning methods (PEPMAN \citep{feng2021machine} and DeePromoter \citep{oubounyt2019deepromoter}).

\vspace{-2mm}
\subsection{What Features does \name Discover?} \label{subsec:discussion_features}
\vspace{-2mm}

The performance gains reported in \Cref{subsec:exp:performance} are driven by the features that \name discovers. For concreteness, \Cref{app:additional_results} shows a tree learnt by \name for the \textit{Pol II pausing} task. Because \name is interpretable by design (each feature includes a human-readable description and executable code) we can directly examine what the model has learned through these semantic representations. This allows us to make the following observations:

\noindent\textbf{Obs.~1---DEFT crosses categories.} Across datasets, \name finds not only \emph{composition windows} and \emph{position checks}, but also features related to \emph{layout/spacing} (spacing, block transitions) and \emph{physics / epigenetics} (stacking energy, CpG spacing). These are higher-level abstractions tied to sequence structure, which leads to the following taxonomy:

\begin{mdframed}[style=taxobox]

\noindent\textbf{Position checks}: single/few bases at fixed sites.\\
\emph{e.g.,} \texttt{pos\_50\_is\_G\_and\_pos\_51\_is\_T} (Pol II), 
\texttt{pos\_201\_is\_pyrimidine} (Prom.)

\smallskip
\noindent\textbf{Composition windows}: Percentage of GC/AT or counts over a span (local environment).\\
\emph{e.g.,} \texttt{upstream\_G\_content\_20\_49} (Pol II), 
\texttt{Interrupted\_GC\_Rich\_6mer\_Proportion} (Enh.)

\smallskip
\noindent\textbf{Motifs}: presence of specific $k$-mers (including palindromes).\\
\emph{e.g.,} \texttt{AP1\_Motif\_Variant\_Proportion} (Enh.), 
\texttt{GAAG\_CTTC\_Palindromic\_Proportion} (Enh.)

\smallskip
\noindent\textbf{Layout / spacing}: arrangement grammar (spacing, periodicity, GC\,$\leftrightarrow$\,AT block transitions).\\
\emph{e.g.,} \texttt{A\_Flanked\_CpG\_Transition\_Density} (Prom.), 
\texttt{Boundary\_GC\_AT\_Transition\_Density} (Enh.)

\smallskip
\noindent\textbf{Physics / epigenetics} --- biophysical or CpG-based context proxies.\\
\emph{e.g.,} \texttt{upstream\_base\_pair\_stacking\_energy\_10\_49} (Pol II), 
\texttt{Potential\_Methylation\_Site\_Density} (Prom.)

\smallskip
\end{mdframed}

\smallskip
\noindent\textbf{Obs.~2---Roots are higher-level.} Root splits are typically \emph{composition windows, layout/spacing, physics / epigenetics} (e.g., upstream G content 20--49; GC$\leftrightarrow$AT block transitions), integrating many positions. Deeper nodes then refine with \emph{position checks} (e.g., pos 50 purine / 51 pyrimidine), consistent with progressive specialization along the tree.

\smallskip
\noindent\textbf{Obs.~3---Dataset tendencies.} For the \textit{Pol II pausing} dataset, \name emphasizes \emph{physics and composition} around the pause; in \textit{Promoters}, it highlights \emph{CpG context} and \emph{layout/spacing}; and in \textit{Enhancers}, it surfaces \emph{motifs} (AP-1 variant, palindromes) plus \emph{layout/spacing} over GC/AT blocks. This shows that \name adapts to the characteristics of each dataset.

\vspace{-3mm}
\subsection{\name~ Allows to Control the Balance Between Interpretability and Performance} \label{subsec:tradeoff}

\begin{table}[h]
    \centering
    \caption{\textbf{Performance comparison} at different maximum tree depths, showing training and test accuracies for both methods. We report $\text{mean}_{(\text{std})}$ for 5 seeds.}
    \label{tab:tradeoff_interpretability_accuracy}
    \begin{tabular}{ccccc}
        \toprule
        & \multicolumn{2}{c}{Train} & \multicolumn{2}{c}{Test} \\
        \cmidrule(lr){2-3} \cmidrule(lr){4-5}
        $d$ & \texttt{DEFT} & $\texttt{DEFT}_{\texttt{perf}}$ & \texttt{DEFT} & $\texttt{DEFT}_{\texttt{perf}}$ \\
        \midrule
        1 & ${\small 0.769}_{\scriptscriptstyle (0.004)}$ & ${\small 0.820}_{\scriptscriptstyle (0.012)}$ & ${\small 0.762}_{\scriptscriptstyle (0.004)}$ & ${\small 0.810}_{\scriptscriptstyle (0.012)}$ \\
        2 & ${\small 0.836}_{\scriptscriptstyle (0.017)}$ & ${\small 0.872}_{\scriptscriptstyle (0.020)}$ & ${\small 0.823}_{\scriptscriptstyle (0.024)}$ & ${\small 0.855}_{\scriptscriptstyle (0.014)}$ \\
        3 & ${\small 0.872}_{\scriptscriptstyle (0.009)}$ & ${\small 0.893}_{\scriptscriptstyle (0.006)}$ & ${\small 0.856}_{\scriptscriptstyle (0.014)}$ & ${\small 0.867}_{\scriptscriptstyle (0.007)}$ \\
        \bottomrule
    \end{tabular}
\end{table}

\begin{table}[h]
\centering
\caption{\textbf{Halstead metrics}: Volume, effort and difficulty (lower is better). Removing the intepretability constraints leads to more complex features.}
\begin{tabular}{lccc}
\toprule
Model & Hal. Vol. & Hal. Effort  & Hal. Dif. \\
\hline
\name     & 15.5 & 15.5 & 0.875 \\
$\texttt{DEFT}_{\texttt{perf}}$ & 19.7 & 17.3 & 1.0   \\
\bottomrule
\end{tabular}

\label{tab:halstead}
\end{table}

\textbf{Methodology.} While the previous results demonstrate that \name~ generates high-level features that are discriminative (hence achieving both \textit{interpretability} and \textit{predictive} accuracy), we now show that \name also enables practitioners to control the balance between these objectives according to their requirements. To demonstrate this, we encourage the exploration and exploitation of the search space and remove the interpretability constraints from the reflection prompts in $\mathcal{I}$. These require that features be simple (the feature should be easy to understand and must not combine multiple complex phenomena), intuitive (the computational logic should be immediately clear to a biologist), and  relevant (the feature must be relevant to the biological prediction task). This modification allows \name~ to focus solely on generating features that maximize predictive performance, regardless of complexity. We denote this  as $\texttt{DEFT}_{\texttt{perf}}$.

\textbf{Results.} We report the training and test accuracies for small tree depths in \cref{tab:tradeoff_interpretability_accuracy} (\textit{Pol II pausing} task).
While the gap between $\texttt{DEFT}_{\texttt{perf}}$ and $\texttt{DEFT}$ reduces as $d$ increases, there is a strong performance difference at depth $1$. To provide intuition for this observation, we show in \cref{app:additional_results} an example of a feature discovered by $\texttt{DEFT}_{\texttt{perf}}$ at the root node, which exploits composite sequence patterns. 
We also compute Halstead complexity metrics \citep{halstead1977elements} for the features discovered by  \texttt{DEFT} and $\texttt{DEFT}_{\texttt{perf}}$. The median results across the generated features are presented in \Cref{tab:halstead} and confirm our intuition that removing the interpretability constraints leads to more complex features.

\subsection{Ablations} \label{subsec:exp:ablations}

\textbf{Methodology.} Having previously demonstrated that \name~ can explore efficiently the space of possible features to generate predictive features, we now ablate its different components to verify their contribution to the overall search performance.
We conduct ablation experiments by removing the following key components:
$\blacktriangleright$~ \textbf{Prior knowledge}:  $\name_{\texttt{no prior}}$ removes the semantic information from the prompts for both population initialization and the reflection mechanism. This includes replacing the description of the task with generic information.
$\blacktriangleright$~ \textbf{Adaptivity}: $\name_{\texttt{no adapt}}$ fits a \texttt{CART} model on a feature set consisting of the raw features and the features generated for the root node (which corresponds to the whole dataset). This contrasts \name~ which dynamically generates features at each node during tree induction and hence considers the local data characteristics. $\blacktriangleright$~\textbf{Reflection.}  $\name_{\texttt{no ref}}$ removes the reflection mechanism, relying solely on the LLM for population initialization.

\begin{figure}[t]
    \centering
    \includegraphics[width=0.7\linewidth]{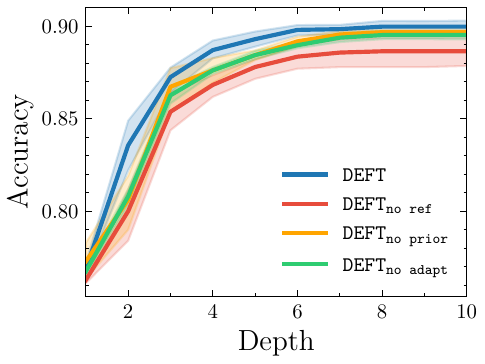}
    \caption{\textbf{Ablations.} Mean and confidence intervals at $95\%$ for $5$ seeds.}
    \label{fig:ablations}
    \vspace{-3mm}
\end{figure}

\textbf{Results.} We report the search efficiency (training accuracy) for each of these ablations in \cref{fig:ablations} (Pol II pausing), observing that these three components are necessary to achieve optimal search efficiency, in particular at smaller tree depths. Notably, the \textit{reflection mechanism} (which leverages the LLM's in-context learning capabilities) plays a crucial role in navigating the search space of possible feature maps, also explaining the competitive performance of $\name_{\texttt{no prior}}$. We provide detailed analysis of the reflection's performance improvements in \cref{app:additional_results}.

\section{DISCUSSION} \label{sec:discussion}
In this work, we introduced \name, an interpretable tree-based model for DNA sequence analysis. In contrast to traditional decision trees which operate on raw features, \name~ automatically discovers high-level sequence features during tree induction. It leverages LLMs to navigate the search space of possible feature maps, exploiting both their prior knowledge and in-context learning abilities. Through comprehensive experiments on diverse genomic tasks, we demonstrate that \name~ discovers high-level and highly predictive features.

\textbf{Limitations.} As with traditional tree-based approaches, training data variations can influence the resulting tree --- a characteristic inherent to decision tree induction that can be addressed through ensemble methods (e.g., bagging), although we note this comes at the cost of the interpretability of a single tree.  Additionally, \name~ incurs higher computational costs per node compared to conventional trees due to LLM inference during feature generation (see \cref{app:experimental_details}). 

\textbf{Future work.} Future work may investigate the performance of \name~ on datasets with higher dimensionality or other domains with structured modalities (e.g. amino acids). Furthermore, while \name~ demonstrates strong empirical performance, establishing formal guarantees for the convergence of the evolutionary reflection mechanism constitutes an interesting research direction.

\subsubsection*{Acknowledgements}
The authors would like to thank Tennison Liu and the four anonymous AISTATS reviewers for useful comments on an earlier version of the manuscript. NH is funded by Illumina, KK by Roche. This work was supported by Azure sponsorship
credits granted by Microsoft’s AI for Good Research Lab.

\newpage

\bibliographystyle{apalike}
\bibliography{bibliography} 


\clearpage
\appendix
\thispagestyle{empty}

\onecolumn
\aistatstitle{
Supplementary Material}

\section{EXTENDED RELATED WORKS} \label[appendix]{app:extended_related_works}
Our work intersects with several areas in machine learning and computational biology, which we detail below.

\textbf{Machine Learning for DNA Sequence Analysis.} The analysis of DNA sequences using machine learning has seen significant advances in recent years. Traditional approaches rely on position weight matrices and k-mer based models \citep{stormo2000dna}, which provide interpretable results but often lack the expressivity to capture complex sequence patterns. More recently, deep learning architectures have demonstrated superior predictive performance across various genomic tasks. Convolutional neural networks have proven particularly effective for DNA sequence analysis, with \citep{alipanahi2015predicting} showing their ability to learn regulatory motifs and predict transcription factor binding sites. This line of work has been extended through architectures like DeepSEA \citep{zhou2015predicting}, which can identify sequence patterns at multiple spatial scales.
Transformer-based models have further advanced the field, with works like Enformer \citep{avsec2021effective}  or the Nucleotide Transformer \citep{dalla2025nucleotide} demonstrating the ability to capture dependencies in genomic sequences. While these deep learning approaches achieve remarkable performance, their black-box nature and their large scale ($500M$ parameters for the Nucleotide Transformer)  makes their predictions non transparent, which does not answer our original question: \textit{How do we design models that are inherently transparent, yet expressive for DNA sequence analysis?}

\textbf{Interpreting Black Boxes in DNA Sequence Analysis.}
The need for interpretability in genomic applications has led to various approaches to explain machine learning models. Methods like integrated gradients \citep{sundararajan2017axiomatic}, DeepLIFT \citep{shrikumar2017learning}, visualization of attention maps \citep{avsec2021base}, have been widely used to probe these black boxes. 
However, these post-hoc interpretation methods have several limitations. They explain individual predictions rather than providing global model understanding, and the extracted patterns may not faithfully represent the model's predictions \citep{rudin2019stop}. In contrast, \name~ is inherently \textit{transparent}, using a tree structure with human-understandable features.

\textbf{Decision Trees.} Decision trees are valued for their interpretability and ability to capture nonlinear patterns.
Greedy algorithms sequentially grow trees with a top-down approach. Popular methods in this class of algorithms are \texttt{CART} \citep{breiman1984classification}, \texttt{ID3} \citep{quinlan1986induction} and \texttt{C4.5} \citep{quinlan2014c4}. Another branch of methods use combinatorial optimization techniques to search for sparse, optimal trees, e.g.~branch and bound \citep{lin2020generalized} and dynamic programming \citep{aglin2020learning}. Notable works include \texttt{BinOCT} \citep{verwer2019learning}, \texttt{DL85} \citep{aglin2020learning}, \texttt{OSDT} \citep{hu2019optimal}, and \texttt{GOSDT} \citep{lin2020generalized}.
A common limitation of these methods is their reliance on single-feature splits. Hence some works have explored ways to enhance tree expressivity, for example through oblique splits \citep{murthy1994system}. However, these methods explore a restricted search space (e.g. linear combinations of features), and are very difficult to optimize given the non-differentiability of the objective function, which is exacerbated by the high dimensionality of genomic datasets. 

\textbf{LLMs in Scientific Applications.} Recent work has demonstrated the potential of large language models (LLMs) in scientific applications beyond natural language processing. LLMs have shown strong capabilities in tasks like mathematical reasoning \citep{lewkowycz2022solving}, code generation \citep{chen2021evaluating}. Particularly relevant to our work are approaches using LLMs for hypothesis generation in scientific discovery \citep{wang2023hypothesis} and genetic algorithms \citep{liu2024evolution}. The use of LLMs for feature engineering is an emerging area, with recent works exploring their potential for tabular data \citep{hanlarge, hollmann2024large}. However, these approaches typically treat feature generation as a standalone preprocessing step, while \name~ integrates feature discovery in the tree induction process, allowing for adaptivity based on local data characteristics. Furthermore, while \citep{hanlarge, hollmann2024large} mostly construct features based on compositions of simple arithmetic operations (e.g. $+$, $-$, $\times$) applied to continuous features, \name~ can discover high-level features which take into account the sequential nature of the data (motifs, counts...).

\section{EXPERIMENTAL DETAILS} \label[appendix]{app:experimental_details}
\textbf{Code Release.} Code can be found at \url{https://github.com/nicolashuynh/deft}.

\textbf{Compute Resources.} All the experiments were conducted on a machine equipped with a 18-Core Intel Core i9-10980XE CPU, and a NVIDIA GeForce RTX 3080.

\subsection{Details on the Datasets} \label{app:details_dataset}

\textbf{Data Processing.}  \name~ can operate on the original raw features without any preprocessing, since it generates code representations that can take as input dataframes. However, \texttt{CART} can only process continuous or ordinal features. Therefore, for \texttt{CART} we one-hot encode each position in the sequences, yielding a total of $404 = 4\times101$ features. This approach outperformed the alternative of treating nucleotides as ordinal variables (computing the ordering based on the label proportions \citep{hastie2009elements}), which is why we adopt it in \cref{sec:experiments}.

\subsubsection{Pol II Pausing}

The Polymerase II pausing task is based on a dataset \citep{fong2022pausing} publicly available on the GEO platform \citep{GEO}, with accession number \texttt{GSE202749} (licensed under a Creative Commons Attribution 4.0 International License).

\textbf{Background on Pol II Pausing.} We first give some background information on RNA polymerase II pausing. 
RNA Polymerase II (Pol II) is responsible for transcribing protein-coding and many non-coding genes. Its transcription cycle includes an elongation phase where the RNA chain grows \citep{Cramer2019}. This growth is not continuous; Pol II intrinsically pauses at specific sites, temporarily halting nucleotide addition. These pauses are fundamental to gene regulation, as the average transcription velocity---varying from $\leq 0.5$ to $\geq 5$ kb/min \citep{NoeGonzalez2021, Jonkers2015}---is primarily governed by pause frequency and duration. This speed, in turn, kinetically couples to co-transcriptional events such as RNA splicing, polyadenylation, and chromatin modifications \citep{Bentley2014}. The importance of pausing is underscored by the universal conservation of Spt5/NusG, a protein dedicated to its regulation \citep{Mooney2025}. While methods like NET-seq have precisely mapped reproducible pause sites, the full set of sequence determinants remains unknown. Pausing likely involves extensive contacts within the transcription elongation complex (Pol II, DNA, RNA, elongation factors), as well as influences from RNA secondary structure, DNA sequence-dependent nucleosome positioning, and the stability of RNA-DNA hybrids and dsDNA \citep{Veloso2014, Zhang2016}.

\textbf{Description.} Following the description of \citep{fong2022pausing}, this dataset was collected using an enhanced version of NET-seq (eNET-seq), which maps RNA Polymerase II pausing at single-base resolution in human HCT116 cells. The standard NET-seq protocol was modified with optimized MNase digestion, decapping enzymes, and unique molecular identifiers (UMIs) for accurate quantification. The resulting data captures the precise positions of paused RNA Polymerase II complexes across the genome by sequencing the $3'$ ends of nascent RNA transcripts that are protected from MNase digestion by the polymerase. Pause sites in the dataset are defined using three specific quantitative criteria: each pause site must have an eNET-seq signal that exceeds the mean signal of its surrounding $200$ bp window by more than $3$ standard deviations, contain at least 5 reads at the precise pause position, and have a minimum of $5$ additional reads within the surrounding window.

\textbf{Dataset Statistics.}  The DNA sequences have length $101$, and the labels are defined with respect to the central position in each sequence (indexed as position $50$). We summarize the statistics of the dataset in \cref{tab:dataset_statistics}.
The random seeds in the experiments on the Pol II dataset control the training set, while the test set remains fixed.  
\begin{table}[H]
    \centering
    \caption{Pol II pausing dataset characteristics}
    \begin{tabular}{ll}
        \toprule
        \textbf{Characteristic} & \textbf{Value} \\
        \midrule
        Total number of samples & $6000$ \\
        Training set size & $4000$ \\
        Test set size & $2000$ \\
        Dimensionality of the sequences &  $101$\\
        Label distribution (training set) &  Pausing: $51\%$, Non-pausing: $49\%$\\
        \bottomrule
    \end{tabular}
    \label{tab:dataset_statistics}
\end{table}

\subsubsection{Human non-TATA Promoters}

The identification of human non-TATA promoters task uses the dataset from the benchmarks suite \citep{grevsova2023genomic} under the Apache-2.0 license. It involves classifying $251$bp genomic sequences as either non-TATA promoters (positive class; sequences spanning -$200$bp to +$50$bp around a transcription start site lacking a TATA-box) or non-promoter regions (negative class; random $251$bp fragments from human gene regions located after the first exons).

Intuitively, this promoter classification task differs from the Pol II pausing task in several key aspects. The Pol II pausing task focuses on identifying relatively short, localized sequence motifs (within a ~100bp window). In contrast, the non-TATA promoter task requires the features to model a broader, more diverse set of distributed sequence elements  in the absence of the strong, canonical TATA-box signal. 

We give the characteristics of the dataset in \cref{tab:promoters_statistics}.
\begin{table}[H]
    \centering
    \caption{Non-TATA Promoters dataset characteristics}
    \begin{tabular}{ll}
        \toprule
        \textbf{Characteristic} & \textbf{Value} \\
        \midrule
        Total number of samples & $36131$ \\
        Training set size & $27097$ \\
        Test set size & $9034$ \\
        Dimensionality of the sequences &  $251$\\
        Label distribution (training set) &  1: $54\%$, 0: $46\%$\\
        \bottomrule
    \end{tabular}
    \label{tab:promoters_statistics}
\end{table}

\subsubsection{Enhancers}
The enhancer classification task uses MPRA data from \citep{ernst2016genome} (under MIT license). We use the data corresponding to the K562 cell line, minimal promoter, and average replicate. We define the binary labels via a thresholding approach, where positive samples are such that their activity is two standard deviations above the mean (calculated from the training set), and negative samples are such that their activity is two standard deviations below the mean. 
We also subsample the dataset to make the label distribution balanced. The characteristics of the data can be found in \Cref{tab:enhancers_statistics}.

\begin{table}[H]
    \centering
    \caption{Enhancers dataset characteristics}
    \begin{tabular}{ll}
        \toprule
        \textbf{Characteristic} & \textbf{Value} \\
        \midrule
        Total number of samples & $16390$ \\
        Training set size & $16108$ \\
        Test set size & $282$ \\
        Dimensionality of the sequences &  $145$\\
        Label distribution &  1: $50\%$, 0: $50\%$\\
        \bottomrule
    \end{tabular}
    \label{tab:enhancers_statistics}
\end{table}

\subsection{Details on the Method} \label{app:details_method}

\subsubsection{Algorithms: Feature Generation and Inference}
\cref{alg:feature_generation} describes the \textit{feature generation process} at any node $v$, while \cref{alg:prediction} describes how to get predictions from the trees at \textit{inference time}.

\begin{algorithm}[H] 
\caption{Feature generation at node $v$}
\begin{algorithmic}[1] \label{alg:feature_generation}
\REQUIRE Node $v$, tree $\T$, subset $\D$, task context $\C$, population size $M$, number of iterations $K$
\STATE $\mathcal{S}_{v, \T}^{nl} \leftarrow$ serialized sequence of splitting conditions from root to $v$
\STATE $P \leftarrow \emptyset$ \COMMENT{Initial population}
\FOR{$j = 1$ to $M$}
    \STATE $\mathbf{z}_{j} \sim \texttt{LLM}(\cdot \vert \mathcal{S}_{v,\T}^{nl}, \C, \mathcal{I}_{sem})$ \COMMENT{Semantic rep.}
    \STATE $f_j \sim \texttt{LLM}(\cdot \vert \mathbf{z}_{j}, \mathcal{I}_{code})$ \COMMENT{Executable code}
    \STATE $P \leftarrow P \cup \{(f_j, \mathbf{z}_{j})\}$
\ENDFOR
\STATE $P \leftarrow P \cup \{ (\phi_{i}, \bar{\mathbf{z}}_{i}) \}_{i=1}^{d}$ \COMMENT{Add raw features}
\STATE Compute scores $\eta = \min_{\tau \in \mathbb{R}} s(f,\tau,\D)$ for all $f$ in $P$
\FOR{$k = 1$ to $K$}
    \STATE Create few-shot prompt $\mathcal{P}^{nl}$ from $P$ and scores
    \FOR{$\mathcal{I}_{ref} \in \mathcal{I}$}
    \FOR{$m=1$ to $M$}
    \STATE $\mathbf{z}'_{m} \sim \texttt{LLM}(\cdot \vert \mathcal{P}^{nl}, \mathcal{S}_{v,\T}^{nl}, \C, \mathcal{I}_{ref})$
    \STATE $f'_m \sim \texttt{LLM}(\cdot \vert \mathbf{z}'_{m}, \mathcal{I}_{code})$
    \ENDFOR
   \STATE  $P' \leftarrow P' \cup \{ ( f'_{m}, \mathbf{z}'_{m} )\}_{m=1}^M$
   \STATE Compute scores for all features in $P'$
   \ENDFOR
    \STATE  $P \leftarrow \texttt{Top-M}(P\cup P')$  \COMMENT{Selection on scores}
\ENDFOR
\STATE \textbf{return} $(f^*, \mathbf{z}^*, \tau^*) = \arg\min_{(f,\mathbf{z}) \in P, \tau} s(f,\tau,\D)$
\end{algorithmic}
\end{algorithm}

\begin{algorithm}[H]
\caption{Prediction with a fitted tree}
\begin{algorithmic}[1] \label{alg:prediction}
\REQUIRE Fitted tree $t$, input sample $\mathbf{x} \in \mathcal{X}$
\STATE $v \leftarrow \text{root node of } t$
\WHILE{$v$ is not a leaf node}
    \STATE Let $(f_v, \tau_v)$ be the feature and threshold associated with node $v$.
    \STATE $f_v(\mathbf{x}) \leftarrow \text{Compute feature value for sample } \mathbf{x} \text{ using executable code of } f_v$.
    \IF{$f_v(\mathbf{x}) \leq \tau_v$}
        \STATE $v \leftarrow \text{left child of } v$
    \ELSE
        \STATE $v \leftarrow \text{right child of } v$
    \ENDIF
\ENDWHILE
\STATE Let $\hat{y}_v$ be the prediction associated with leaf node $v$.
\STATE \textbf{return} $\hat{y}_v$
\end{algorithmic}
\end{algorithm}

\subsubsection{LLM Hyperparameters}
We detail in \cref{tab:llm_parameters} the hyperparameters of the LLM used throughout our experiments.

\begin{table}[htbp]
    \centering
    \caption{LLM parameters}
    \begin{tabular}{ll}
        \toprule
        \textbf{Parameter} & \textbf{Value} \\
        \midrule
        Model name &  \texttt{gpt-4o} \\
        Version & \texttt{1001} \\
        Temperature & $1$ \\
        Top p & $0.95$ \\
        \bottomrule
    \end{tabular}
    \label{tab:llm_parameters}
\end{table}

\name~ uses a rejection mechanism where invalid features (e.g. which cannot be automatically parsed) are discarded.

While our implementation and experiments utilize \texttt{gpt-4o} (version 1001) as the underlying LLM, we note that \name~ is model-agnostic and can be readily adapted to work with any LLM that demonstrates capabilities in natural language understanding and code generation. For instance, we instantiate \name with the \texttt{gpt-oss} model in \Cref{app:gpt_oss}.

\subsubsection{\name-specific Hyperparameters}
In our experimental section, \name~ uses a population size $M = 10$, and a number of reflections $K=20$. As in traditional genetic algorithms, larger populations would lead to a more thorough search \citep{vrajitoru2000large}, however in practice it is limited by the size of the context window of the LLM  and the token limit rate. Furthermore, more reflection steps would also lead to a more thorough exploration of the feature space, but at the cost of a higher number of LLM inferences. This motivates our choice of hyperparameters, which allows a good trade off between these computational considerations and search performance. 

\name~ uses the Gini index as the splitting criterion, defined as:

\begin{equation}
    Q(\mathcal{D}) = 1 - \left(\frac{\sum_{(x,y) \in \mathcal{D}} \mathds{1}(y=0)}{|\mathcal{D}|}\right)^2  - \left(\frac{\sum_{(x,y) \in \mathcal{D}} \mathds{1}(y=1) }{|\mathcal{D}|}\right)^2
\end{equation}

\subsubsection{Prompts}

\textbf{Prompt Structure}. Each prompt for feature generation contains:
\begin{enumerate}
    \item the \textit{task context} $\mathcal{C}$: describes the input space, the label, the characteristics of the dataset, and the tree induction task
    \item the \textit{node context} $\mathcal{S}_{v,\T}^{nl}$: lists the sequence of splitting conditions from the root node to the current node $v$
    \item \textit{interpretability instructions}: prevents composite features combining multiple mechanisms
    \item \textit{task-specific instructions}: for example, instructions for exploration and exploitation in reflection 
\end{enumerate}

In addition to that, the reflection prompts contain in-context features along with their scores.

\textbf{Examples of Prompts.} In what follows, we provide examples of prompts (Pol II pausing task) for population initialization in \cref{lst:population_initialization}, reflection (exploration in \cref{lst:reflection_exploration}, exploitation in \cref{lst:reflection_exploitation}) and code generation in \cref{lst:code_generation}.

\begin{center}
\begin{lstlisting}[language=Python, caption={\textbf{Example prompt for population initialization}~}, label={lst:population_initialization}, numbers=none]
Your goal is to help with the task of growing a decision tree to predict RNA polymerase pausing.  This is a dataset about RNA polymerase pausing. Given a current node that we want to split in the tree, you will  construct a new feature based on the original DNA sequence, such that this new feature  is discriminative for the prediction task of classifying RNA polymerase pausing. The raw  feature is the DNA sequence of length 101 centered on the pause site. Positions from 0 to 49 included correspond to the upstream region. Position 50 corresponds to the site. Positions from 51 to 100 included are in the downstream region. The possible  nucleotide values are A, C, G, T.

The features are: 
raw_sequence: text (average length: 101.0 characters)

 The dataset has 1644 samples.

<Beginning Splitting conditions from root to current node> 
upstream_G_content_20_49 smaller than 0.250 (Calculate the proportion of guanine (G) nucleotides in the upstream region from positions 20 to 49. Count the number of G nucleotides in this region and divide by 30 to get the proportion.)
<End of Splitting conditions from root to current node>

Leverage your biology expertise and your creativity to generate a good feature (name,  description, justification). This feature can be simple. Importantly, take into account  the contextual history given above, which defines the sequence of splitting conditions  from the root node up to the current node. Another important  point: The feature should be interpretable. The feature should capture a single, clear  biological mechanism. Use only basic sequence properties that a biologist could understand  and find intuitive. Avoid combining multiple biological mechanisms into one feature,  which would introduce complexity. Be very explicit in your description on how you compute  the feature. Return the feature in the following JSON format:
{"rationale": "rationale", "description": "your_feature_description", "name": "your_feature_name"}
Only return the JSON object, with no additional text. 
\end{lstlisting}
\end{center}

\newpage

\begin{lstlisting}[language=Python, caption={\textbf{Example prompt for reflection (exploration)}~}, label={lst:reflection_exploration}, numbers=none]

Your goal is to help with the task of growing a decision tree to predict RNA polymerase pausing.  This is a dataset about RNA polymerase pausing. Given a current node that we want to split in the tree, you will  construct a new feature based on the original DNA sequence, such that this new feature  is discriminative for the prediction task of classifying RNA polymerase pausing. The raw  feature is the DNA sequence of length 101 centered on the pause site. Positions from 0 to 49 included correspond to the upstream region. Position 50 corresponds to the site. Positions from 51 to 100 included are in the downstream region. The possible  nucleotide values are A, C, G, T.

The features are:
raw_sequence: text (average length: 101.0 characters)

 The dataset has 1644 samples.

Given a current node that we want to split in the tree, you will construct a new  feature based on the original DNA sequence, such that this new feature is discriminative  for the prediction task of classifying RNA polymerase pausing. The feature should be: 1. Biologically interpretable - based on possible mechanisms of transcription and intuitive  for biologists 2. Computationally clear - specify exact positions and calculation methods,  3. Complementary to previous splitting conditions from the root to current node. I am going to give you an initial population of features with their respective  scores (lower is better). The feature you generate should be as different as possible from the initial population in order to explore new ideas. 

Another important point: The feature should be interpretable. To assess interpretability, please  consider the following aspects: 1) Simplicity: the should be simple and easy to understand.  An interpretable feature should not be overly complex (and not involve multiple complex  phenomena). Hence features which incorporate too many multiple distinct components are not  simple and should not be generated. 2) Intuitiveness: The feature should be intuitive and  easy to explain to a biologist 3) Relevance: the feature should be relevant to the  biological prediction task. 

Return three things: 1. Rationale: describe step by  step how you came up with this feature, taking into account both the splitting conditions  and the population of candidate features. If you feature involves a physical or biological  mechanism, say why it is relevant. 2. Description: precise calculation method 3. Name:  clear and descriptive. Return in the following JSON format:
{"rationale": "your_rationale", "description": "your_feature_description", "name": "your_feature_name"}
Only return the JSON object, with no additional text. Do not include "json" in front of it

<Beginning of the population of features> 
Here is the list of features along with their score: 
Feature 1 
Score: 0.2667 
 Feature name: upstream_GC_content_10_29 
 Feature description: Calculate the proportion of guanine (G) and cytosine (C) nucleotides in the upstream region from positions 10 to 29. Count the number of G and C nucleotides in this region and divide by 20 to get the proportion.
 Feature code: def add_upstream_GC_content_10_29(X):
    def calculate_gc_content(seq):
        upstream_region = seq[10:30]
        gc_count = upstream_region.count('G') + upstream_region.count('C')
        return gc_count / 20
    
    return X['raw_sequence'].apply(calculate_gc_content)
...
Feature 10 
Score: 0.195
 Feature name: pos_50_is_G_and_pos_51_is_T 
 Feature description: Check if position 50 in the raw sequence is G and position 51 is T. Return 1 if both conditions are true, otherwise 0.
 Feature code: def construct_feature(X):
    return X['raw_sequence'].apply(lambda seq: 1 if len(seq) > 51 and seq[50] == 'G' and seq[51] == 'T' else 0)
<End of the population of features>

<Beginning Splitting conditions from root to current node> 
upstream_G_content_20_49 smaller than 0.250 (Calculate the proportion of guanine (G) nucleotides in the upstream region from positions 20 to 49. Count the number of G nucleotides in this region and divide by 30 to get the proportion.)
<End of Splitting conditions from root to current node>
\end{lstlisting}

\begin{lstlisting}[language=Python, caption={\textbf{Example prompt for reflection (exploitation)}~}, label={lst:reflection_exploitation}, numbers=none]
Your goal is to help with the task of growing a decision tree to predict RNA polymerase pausing.  This is a dataset about RNA polymerase pausing. Given a current node that we want to split in the tree, you will  construct a new feature based on the original DNA sequence, such that this new feature  is discriminative for the prediction task of classifying RNA polymerase pausing. The raw  feature is the DNA sequence of length 101 centered on the pause site. Positions from 0 to 49 included correspond to the upstream region. Position 50 corresponds to the site. Positions from 51 to 100 included are in the downstream region. The possible  nucleotide values are A, C, G, T.

The features are:
raw_sequence: text (average length: 101.0 characters)

 The dataset has 1644 samples.

Given a current node that we want to split in the tree, you will construct a new  feature based on the original DNA sequence, such that this new feature is discriminative  for the prediction task of classifying RNA polymerase pausing. The feature should be: 1. Biologically interpretable - based on possible mechanisms of transcription and intuitive  for biologists 2. Computationally clear - specify exact positions and calculation methods,  3. Complementary to previous splitting conditions from the root to current node. I am going to give you an initial population of features with their respective  scores (lower is better). Very important: First identify common ideas of the top performing solutions in the population. Then base your feature on these common ideas and simplify them to get one good feature, but do not simply combine these common ideas (do not just add them for example, that would create too much complexity).
Another important point: The feature should be interpretable. To assess interpretability, please  consider the following aspects: 1) Simplicity: the should be simple and easy to understand.  An interpretable feature should not be overly complex (and not involve multiple complex  phenomena). Hence features which incorporate too many multiple distinct components are not  simple and should not be generated. 2) Intuitiveness: The feature should be intuitive and  easy to explain to a biologist 3) Relevance: the feature should be relevant to the  biological prediction task. 

Return three things: 1. Rationale: describe step by  step how you came up with this feature, taking into account both the splitting conditions  and the population of candidate features. If you feature involves a physical or biological  mechanism, say why it is relevant. 2. Description: precise calculation method 3. Name:  clear and descriptive. Return in the following JSON format:
{"rationale": "your_rationale", "description": "your_feature_description", "name": "your_feature_name"}
Only return the JSON object, with no additional text. Do not include "json" in front of it

<Beginning of the population of features> 
Here is the list of features along with their score: 
Feature 1 
Score: 0.2667 
 Feature name: upstream_GC_content_10_29 
 Feature description: Calculate the proportion of guanine (G) and cytosine (C) nucleotides in the upstream region from positions 10 to 29. Count the number of G and C nucleotides in this region and divide by 20 to get the proportion.
 Feature code: def add_upstream_GC_content_10_29(X):
    def calculate_gc_content(seq):
        upstream_region = seq[10:30]
        gc_count = upstream_region.count('G') + upstream_region.count('C')
        return gc_count / 20
    
    return X['raw_sequence'].apply(calculate_gc_content)
...
Feature 10 
Score: 0.195
 Feature name: pos_50_is_G_and_pos_51_is_T 
 Feature description: Check if position 50 in the raw sequence is G and position 51 is T. Return 1 if both conditions are true, otherwise 0.
 Feature code: def construct_feature(X):
    return X['raw_sequence'].apply(lambda seq: 1 if len(seq) > 51 and seq[50] == 'G' and seq[51] == 'T' else 0)
<End of the population of features>


<Beginning Splitting conditions from root to current node> 
upstream_G_content_20_49 smaller than 0.250 (Calculate the proportion of guanine (G) nucleotides in the upstream region from positions 20 to 49. Count the number of G nucleotides in this region and divide by 30 to get the proportion.)
<End of Splitting conditions from root to current node>
\end{lstlisting}

\begin{lstlisting}[language=Python, caption={\textbf{Example prompt for code generation}~}, label={lst:code_generation}, numbers=none]
Your goal is to generate a python code to construct a feature, based on a dataframe.

 You should build this feature using the following original features:
raw_sequence: text (average length: 101.0 characters)

The feature you should generate has the following characteristics:
Feature name: pos_50_is_G_and_pos_51_is_T 
 Feature description: Check if position 50 in the raw sequence is G and position 51 is T. Return 1 if both conditions are true, otherwise 0.

The raw feature is the DNA sequence of length 101 centered on the pause site. Positions from 0 to 49 included correspond to the upstream region. Position 50 corresponds to the site. Positions from 51 to 100 included are in the downstream region. The possible  nucleotide values are A, C, G, T. Give instantly executable code without example usage. Only return the Python function,  with no additional text. The name of the argument should be 'X'. Only output one function,  this is very important. The function should only return the new feature column. Start your  output with the string 'def {Function Name}(X):' do not include 'python' in front of it

\end{lstlisting}

\subsubsection{Computational Overhead} \label{app:computational_overhead}
Compared to conventional decision tree induction (e.g. \texttt{CART}), \name~ incurs an additional computational overhead which mostly comes from the LLM inference time at each node during tree construction, both for generating the initial population of candidate features and during the reflection mechanism. As such, the cost scales linearly with the number of reflection steps $K$. 
For example, each split for tree induction on the Pol II pausing task took on average $353$ seconds. We note that the overall run time could be greatly reduced by parallelizing the construction of the splits at the same depth, as the total cost would then scale linearly in the depth, rather than on the number of nodes. This constitutes an interesting avenue for future work, but it falls outside the scope of the current work as our key contributions are conceptual and methodological (i.e. a method to embed feature generation into tree induction for DNA sequence analysis).

We emphasize that this computational overhead allows \name~ to discover novel features which are highly discriminative, hence requiring fewer splits to achieve comparable performance to conventional trees.  Hence, the comparison between \texttt{DEFT} and \texttt{CART} must consider two axes: \textit{computational cost} and \textit{search space complexity}. While \texttt{CART}'s search space is highly constrained (it can only evaluate splits along the predefined raw features), \name explores an effectively infinite search space of potential features, and the higher per-node cost is directly exchanged for greater performance (cf \Cref{fig:comparison_with_baselines_accuracy}). Finally, assuming sufficient compute is available for inference, the overall runtime could be significantly reduced through parallelization: since the feature generation process for nodes at the same depth is independent, the tree's branches can be constructed in parallel.

Finally, it is important to note that \name~ provides valuable \textit{insights} for the practitioner by \textit{automatically} discovering \textit{human-interpretable features} that can capture high level sequence patterns, going beyond single-position splits. 

\textbf{Remark on Computational Cost with Respect to Sequence Length.} 
The computational cost of \name~ depends on the complexity of the Python code generated for the features at each split. While the LLM cost for generating the feature descriptions can realistically be assumed to be independent of sequence length (within its context window, and when not providing in-context samples), the cost of executing the generated feature code on a sequence of length $L$ varies. Crucially, \name~ can generate features whose computational complexity scales linearly with $L$ (e.g. counting motifs, calculating GC content in a window). For instance, the "upstream G content 20-49" feature involves a simple string scan over a fixed-size or proportional window, resulting in $O(L)$ or even $O(1)$ complexity if the window size is constant relative to $L$. This contrasts with methods like standard Transformers whose self-attention mechanism often leads to quadratic complexity with sequence length. Therefore, if \name~ predominantly generates features with linear or sub-linear complexity, its application can potentially be significantly more scalable for very long sequences compared to models with inherent quadratic or higher-order scaling with $L$. The practitioner can also guide the LLM (via prompts during the reflection step) to favor computationally simpler features if scalability for long sequences is a primary concern.

\subsection{Details on the Baselines}

\subsubsection{OC1}
\texttt{OC1} \citep{murthy1994system} is a method to build oblique decision trees (\textit{multivariate} trees). Unlike axis-aligned trees that split nodes based on a single feature, \texttt{OC1} creates splits using linear combinations of multiple features (i.e. oblique hyperplanes). It identifies these hyperplanes by first finding the best axis-parallel split and then employing a randomized hill-climbing algorithm. This algorithm iteratively perturbs the hyperplane's coefficients to improve the Gini criterion, and we consider $50$ random restarts to escape local optima.

\subsubsection{k-mers Features}
K-mers are short, fixed-length ($k$) subsequences of DNA. In \cref{subsec:exp:performance}, the input DNA sequences are featurized by counting the occurrences of all possible k-mers (e.g., for $k=2$, "AA", "AC", "AG", "AT", "CA", etc.). These counts then serve as input features to a decision tree. While $k$-mers can capture local sequence motifs without explicit prior biological knowledge, they constitute a form of manual feature engineering. 

\subsubsection{Deep Learning Baselines}
In \Cref{subsec:exp:performance}, we tune the hyperparameters of the CNN and the Transformer models using a grid search over a set of hyperparameters which we show in \Cref{tab:cnn_grid_search} (CNN) and \Cref{tab:transformer_grid_search}    (Transformer). We optimize the validation AUPRC (with a $0.9/0.1$ train/validation split). For model training, we perform early stopping with a patience of $5$, and set the maximum number of epochs to $50$.

\begin{table}[h]
  \centering
  \caption{CNN grid-search hyperparameters and their ranges.}
  \label{tab:cnn_grid_search}
  \begin{tabular}{ll}
    \toprule
    Hyperparameter & Values \\
    \midrule
    Conv1 filters  & \{32, 64\} \\
    Conv1 kernel size  & \{2, 12, 16\} \\
    Conv2 filters & \{64, 128\} \\
    Dense units & \{64, 128\} \\
    Dropout rate & \{0, 0.2, 0.5\} \\
    Learning rate & $\{1\times10^{-3}, 1\times10^{-4}\}$ \\
    Batch size & \{32, 128\} \\
    \bottomrule
  \end{tabular}
\end{table}

\begin{table}[h]
  \centering
  \caption{Transformer grid-search hyperparameters and their ranges.}
  \label{tab:transformer_grid_search}
  \begin{tabular}{ll}
    \toprule
    \textbf{Hyperparameter} & \textbf{Values} \\
    \midrule
    Hidden dimension & \{32, 64, 128\} \\
    Attention heads & \{2, 4\} \\
    Encoder layers & \{1, 2, 4\} \\
    Dropout rate & \{0, 0.2, 0.5\} \\
    Learning rate & $\{1\times 10^{-3}, 1\times 10^{-4}\}$ \\
    Batch size & \{32, 128\} \\
    \bottomrule
  \end{tabular}
\end{table}

\newpage
\section{ADDITIONAL RESULTS} \label[appendix]{app:additional_results}

\begin{table}[H]
\centering
\footnotesize
\caption{\textbf{Summary of additional results.}}
\label{tab:appendix_roadmap}
\setlength{\tabcolsep}{6pt}
\begin{tabular}{@{}p{0.30\linewidth} p{0.48\linewidth} p{0.18\linewidth}@{}}
\toprule
\textbf{Section / Reference} & \textbf{Summary} & \textbf{Results} \\
\midrule
\Cref{app:example_no_interpretability} & Feature found without interpretability constraints. & — \\
\Cref{app:cart_no_reg} & Unregularized \texttt{CART} increasingly overfits as depth grows. & Fig.~\ref{fig:cart_no_reg} \\
\Cref{app:other_metrics} & F1, precision, recall, and AUPRC across datasets. & Fig.~\ref{fig:perf_other_metrics} \\
\Cref{app:nucleotide_transformer} & \name\ vs.\ Nucleotide Transformer (fine tuned with LoRA) for context. & Tab.~\ref{tab:nucleotide_transformer_comparison} \\
\Cref{app:specific_baselines} & Specialized CNN/attention baselines: \texttt{PEPMAN} (Pol II), \texttt{DeePromoter} (Promoters). & Tab.~\ref{tab:pol2-pausing}, Tab.~\ref{tab:promoter-id} \\
\Cref{app:gpt_oss} & \name\ instantiated with an open-source LLM (\texttt{gpt-oss}). & Tab.~\ref{tab:blackbox-comparison}, \\
\Cref{app:ensemble} & Ensembling \name\ trees on the same split boosts Acc./AUPRC. & Tab.~\ref{tab:deft_ensemble} \\
\Cref{app:reflection} & Reflection mechanism analysis. & Fig.~\ref{fig:reflection_scores} \\
\Cref{app:expert_evaluation} & Evaluation of the interpretability of the features &  \\
\Cref{app:dnabert_hyena} & Comparison with two foundation models & Tab.~\ref{tab:performance_comparison_dnabert} \\
\bottomrule
\end{tabular}
\end{table}

\subsection{Example of Feature Without Interpretability Constraints} \label{app:example_no_interpretability}
We provide an example of a feature found by \name~ for the Pol II pausing task when we remove the interpretability constraints from the reflection prompts (see  \cref{subsec:tradeoff} for more details).

This feature illustrates how relaxing interpretability constraints leads to the discovery of composite sequence patterns: it combines position-specific G densities upstream with broader GCT content downstream, achieving higher discriminative power through a more sophisticated feature compared to the ones shown in \Cref{fig:decision_tree}. While the feature is more complex, it is worth noting that the semantic and code representations obtained with \name~ still provide transparency into the feature computation, facilitating both analysis and potential simplification by domain experts.

\begin{mdframed}[style=feature]
\bmag{\underline{Feature name}:} {\texttt{upstream and segmented downstream GCT density}}

\bmag{\underline{Feature description}: } "Calculate the density of G nucleotides in six upstream segments (positions 25-28, 29-33, 34-38, 39-43, 44-47, and 48-49) by dividing the count of G nucleotides by 4, 5, 5, 5, 4, and 2 respectively. At the pause site (position 50), consider the presence of a G nucleotide as 1. Additionally, calculate the density of G, C, and T nucleotides in three downstream segments (positions 51-63, 64-75, and 76-100) by dividing the combined count of G, C, and T nucleotides by 13, 12, and 25 respectively. Combine these densities to get a comprehensive measure."
\end{mdframed}

\subsection{CART Without Complexity Penalty} \label{app:cart_no_reg}

\textbf{Methodology.} In contrast to \cref{subsec:exp:performance}, where we set the minimum number of samples per leaf to $1\%$ of the training set size for \texttt{CART} to prevent overfitting, we now evaluate an unregularized baseline $\texttt{CART}_{\texttt{no reg}}$ with zero minimum samples per leaf.

\begin{figure}[htbp]
    \centering
    \includegraphics[width=0.5\textwidth]{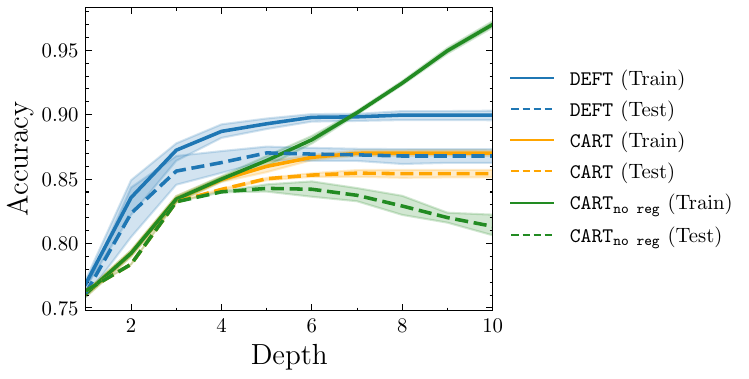}
    \caption{\textbf{CART trees can overfit.} Deep trees constructed with $\texttt{CART}$ overfit the training set when there is no explicit regularization mechanism. We report the mean and confidence intervals at the $95\%$ level for $5$ seeds.}
    \label{fig:cart_no_reg}
\end{figure}

\textbf{Results.} As shown in \cref{fig:cart_no_reg} for the Pol II pausing task, while $\texttt{CART}_{\texttt{no reg}}$ achieves increasingly higher training accuracy at greater tree depths, this comes at the cost of deteriorating test accuracy, an indication of overfitting. 

\subsection{Other Classification Metrics} \label{app:other_metrics}
\begin{figure}[htbp]
\centering

\par\vspace{0.4em}\textbf{Pol II pausing}\par\vspace{0.25em}
\begin{subfigure}[b]{0.24\textwidth}
    \includegraphics[width=\textwidth]{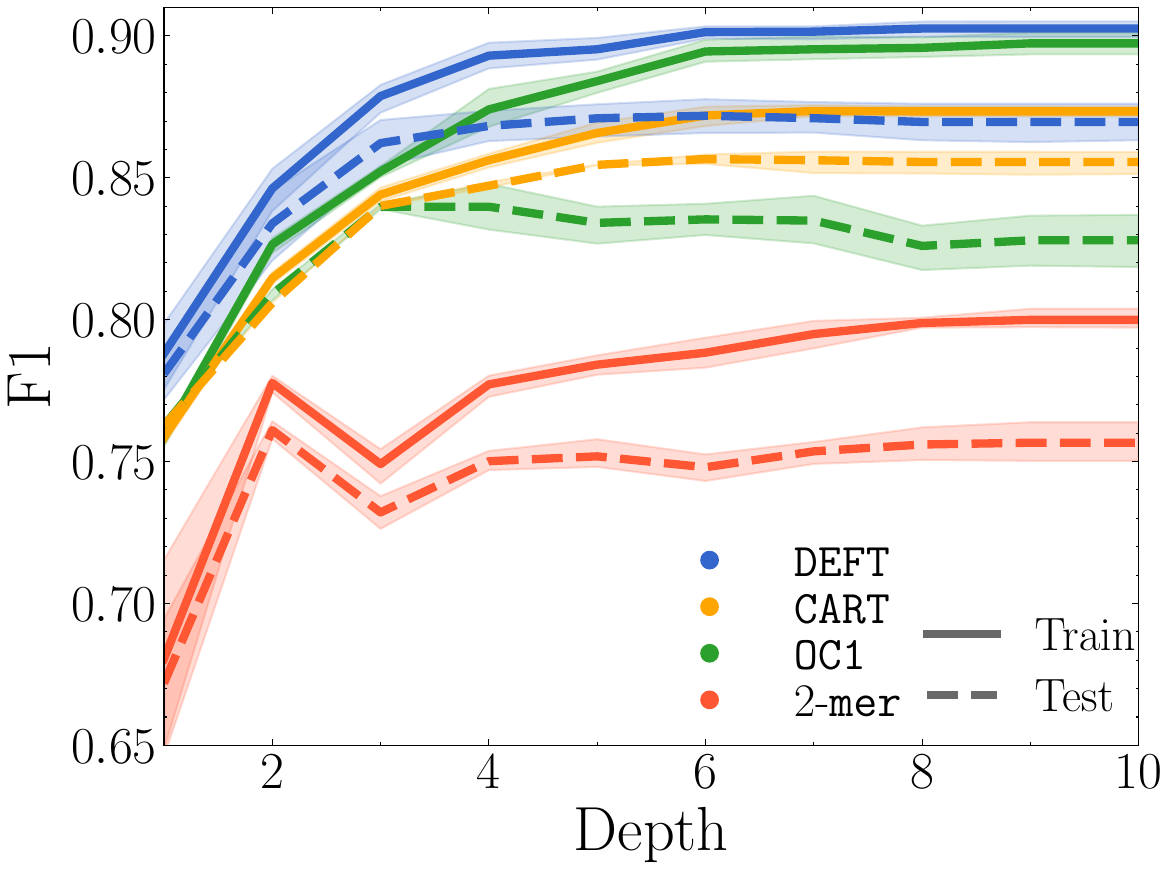}
    \caption{F1 score}
    \label{fig:dsA_f1}
\end{subfigure}
\hfill
\begin{subfigure}[b]{0.24\textwidth}
    \includegraphics[width=\textwidth]{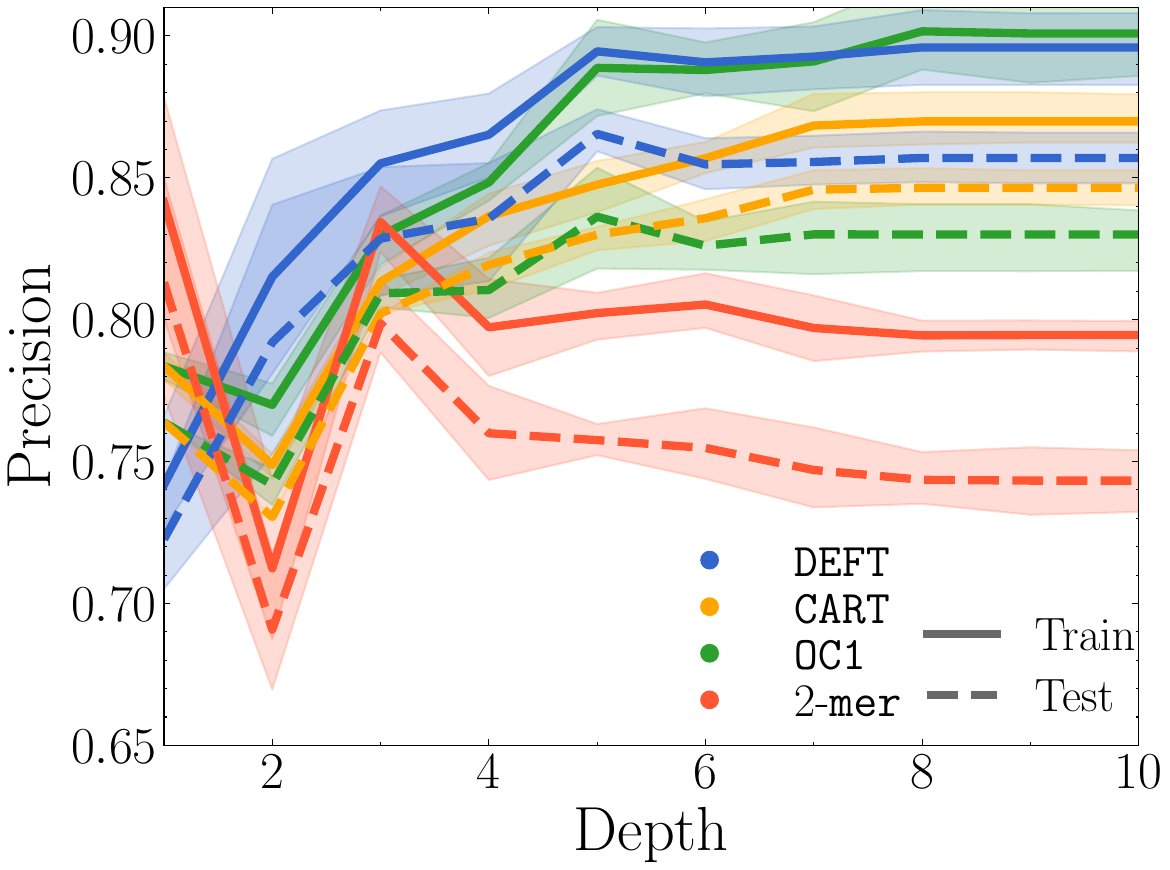}
    \caption{Precision}
    \label{fig:dsA_precision}
\end{subfigure}
\hfill
\begin{subfigure}[b]{0.24\textwidth}
    \includegraphics[width=\textwidth]{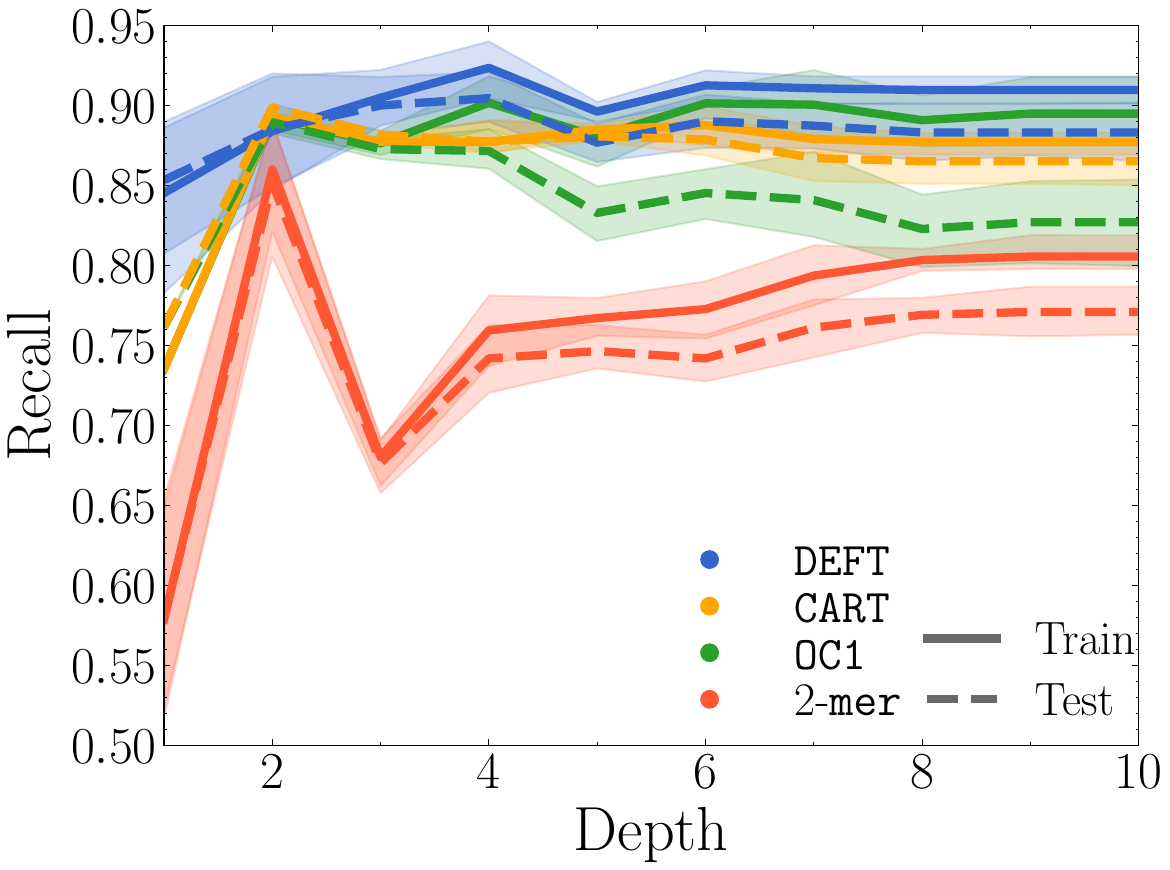}
    \caption{Recall}
    \label{fig:dsA_recall}
\end{subfigure}
\hfill
\begin{subfigure}[b]{0.24\textwidth}
    \includegraphics[width=\textwidth]{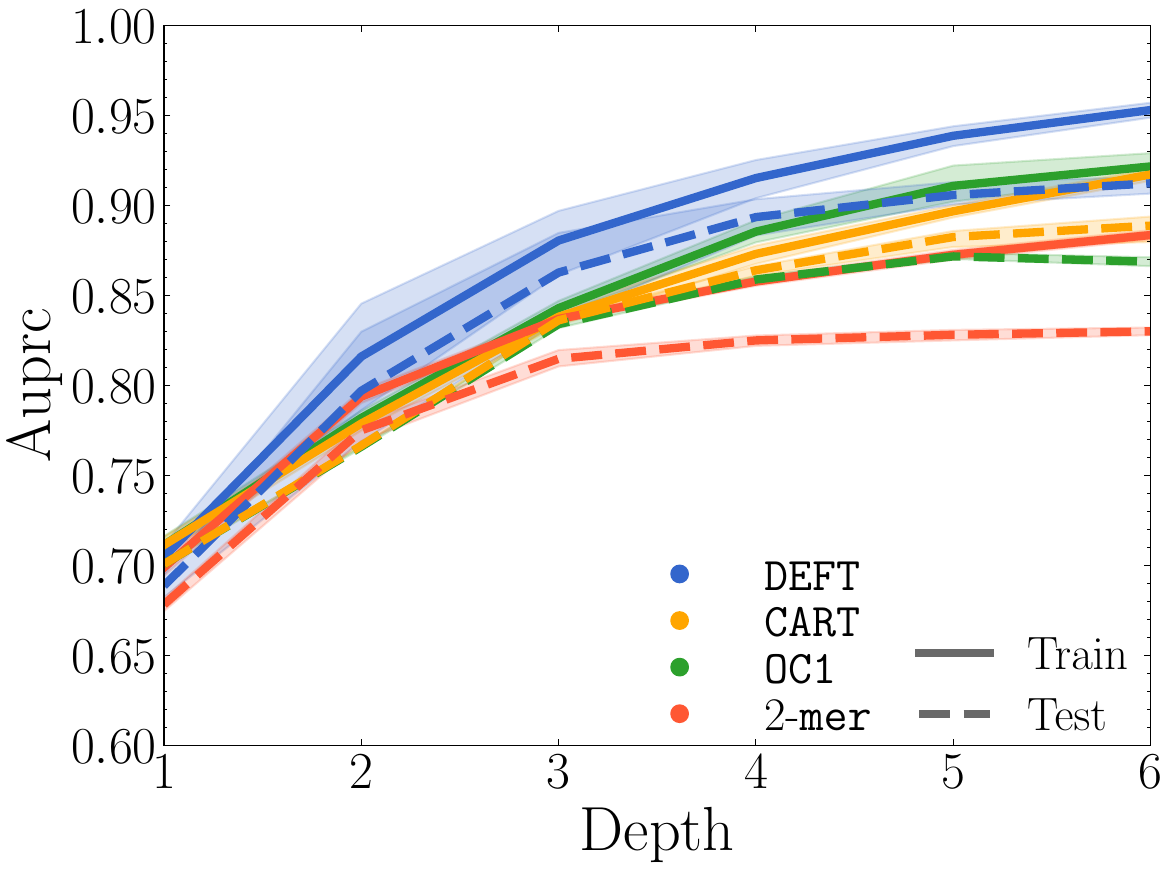}
    \caption{AUPRC}
    \label{fig:dsA_auprc}
\end{subfigure}

\par\vspace{0.8em}\textbf{Promoters}\par\vspace{0.25em}
\begin{subfigure}[b]{0.24\textwidth}
    \includegraphics[width=\textwidth]{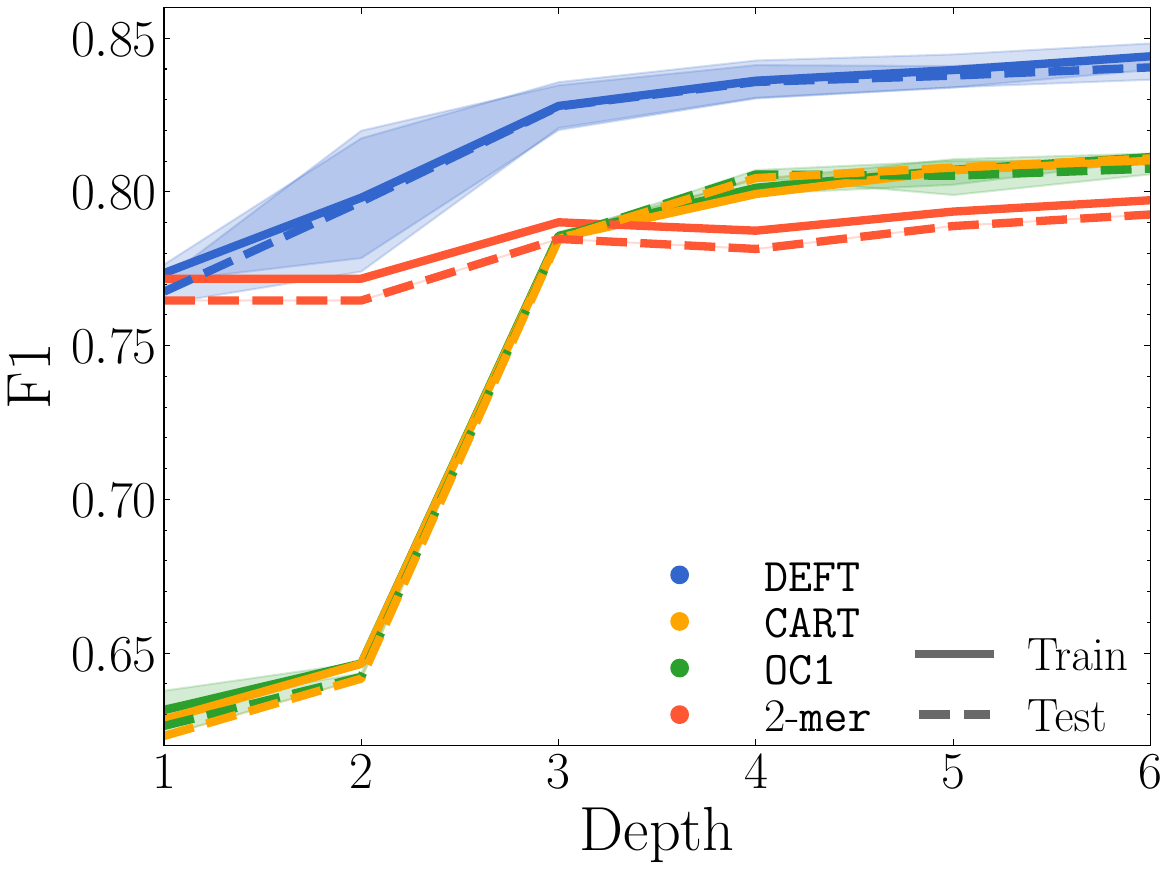}
    \caption{F1 score}
    \label{fig:dsB_f1}
\end{subfigure}
\hfill
\begin{subfigure}[b]{0.24\textwidth}
    \includegraphics[width=\textwidth]{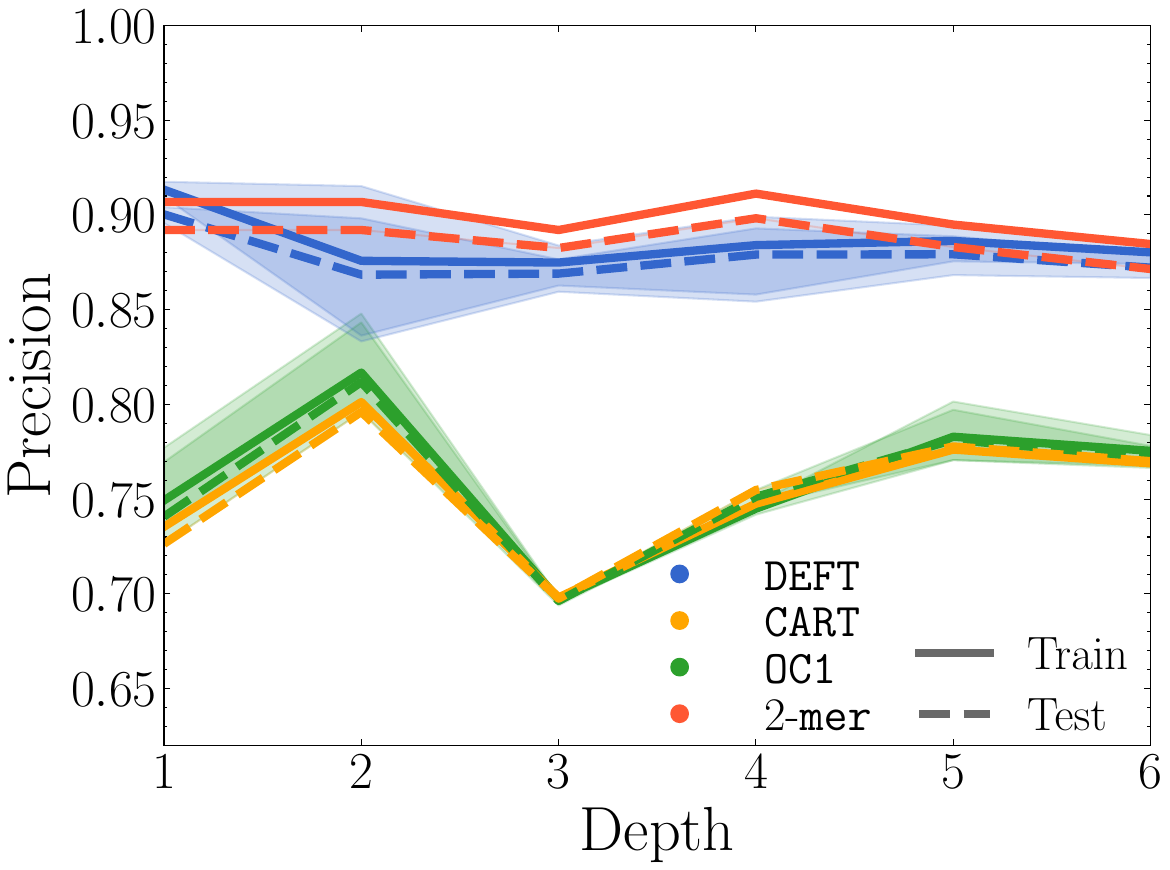}
    \caption{Precision}
    \label{fig:dsB_precision}
\end{subfigure}
\hfill
\begin{subfigure}[b]{0.24\textwidth}
    \includegraphics[width=\textwidth]{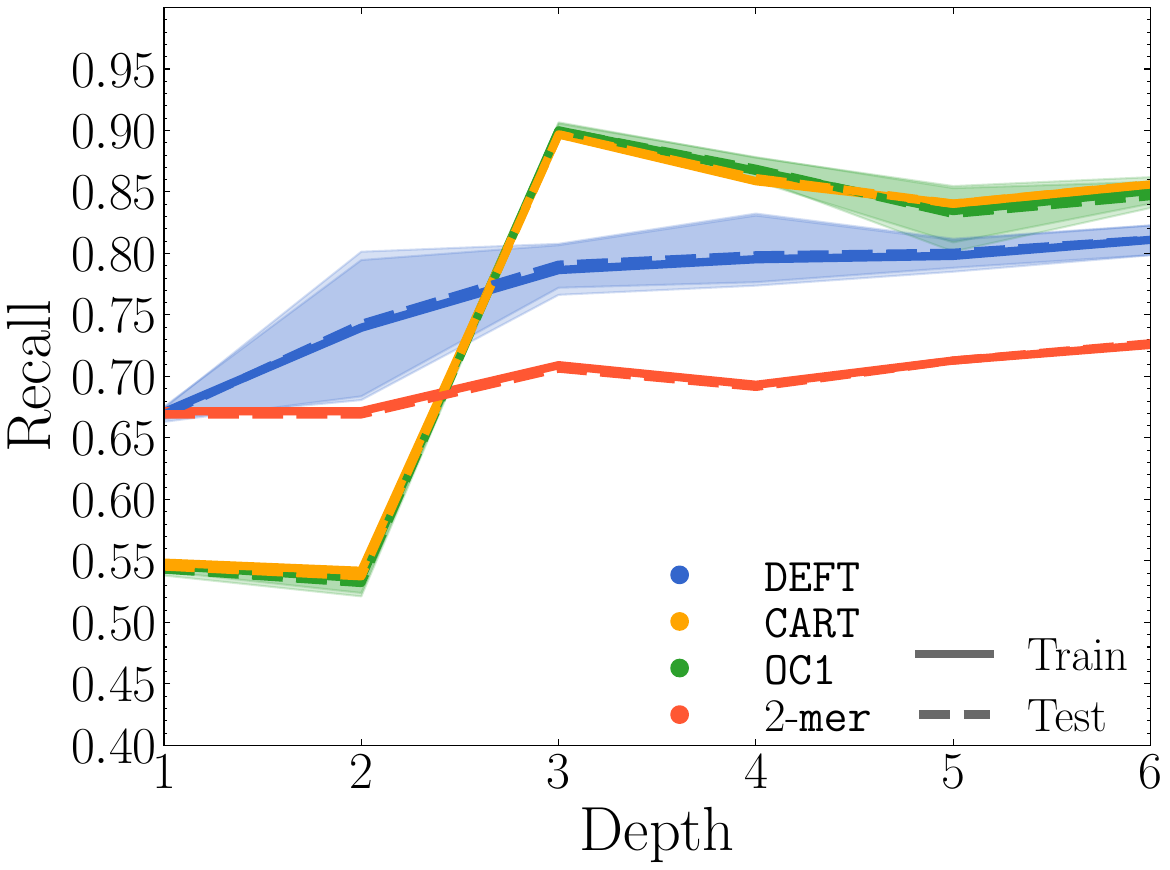}
    \caption{Recall}
    \label{fig:dsB_recall}
\end{subfigure}
\hfill
\begin{subfigure}[b]{0.24\textwidth}
    \includegraphics[width=\textwidth]{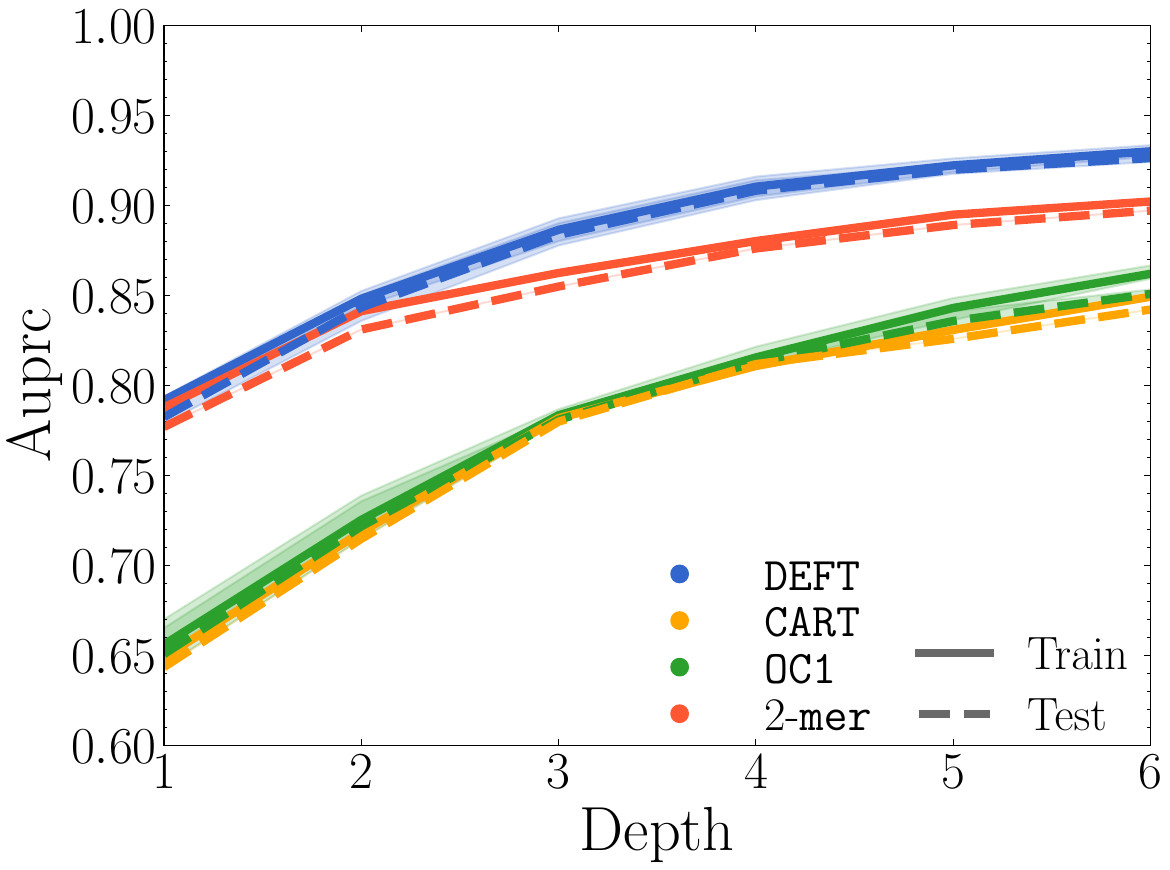}
    \caption{AUPRC}
    \label{fig:dsB_auprc}
\end{subfigure}

\par\vspace{0.8em}\textbf{Enhancers}\par\vspace{0.25em}
\begin{subfigure}[b]{0.24\textwidth}
    \includegraphics[width=\textwidth]{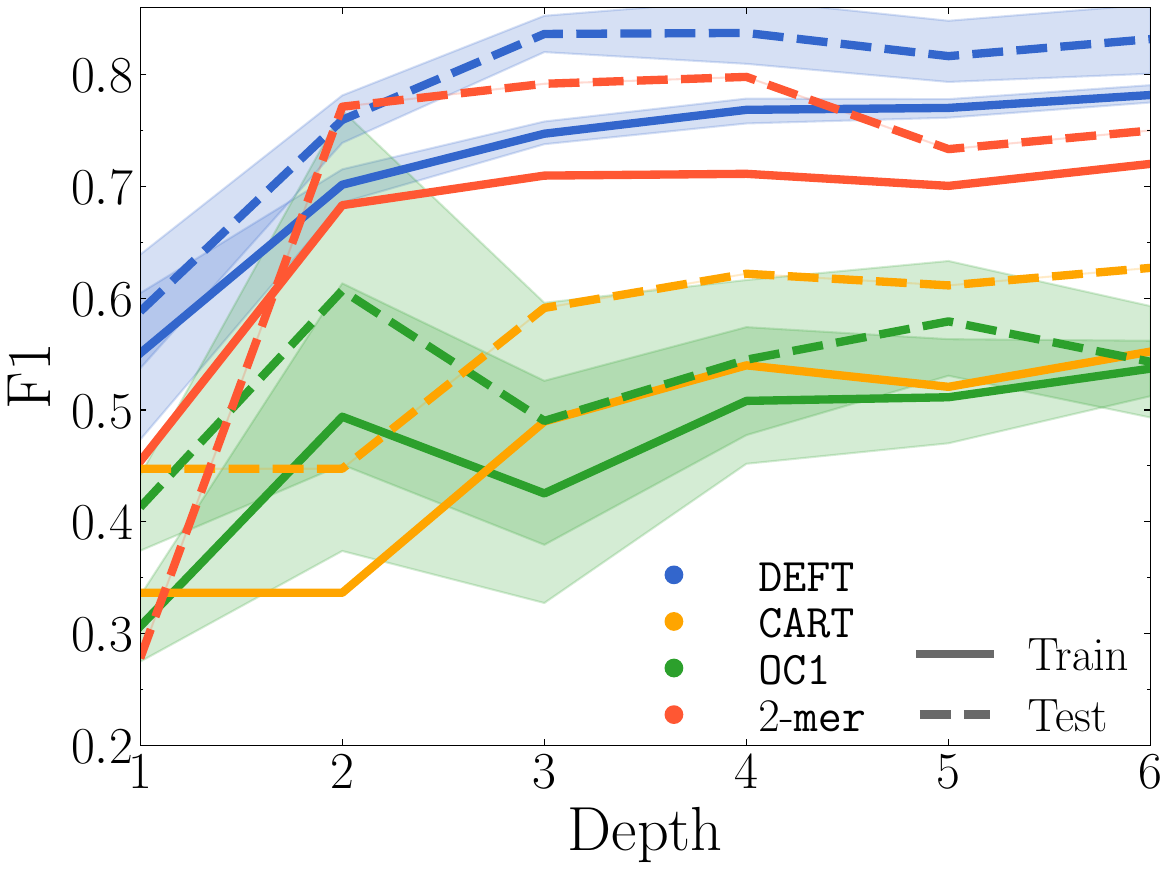}
    \caption{F1 score}
    \label{fig:dsC_f1}
\end{subfigure}
\hfill
\begin{subfigure}[b]{0.24\textwidth}
    \includegraphics[width=\textwidth]{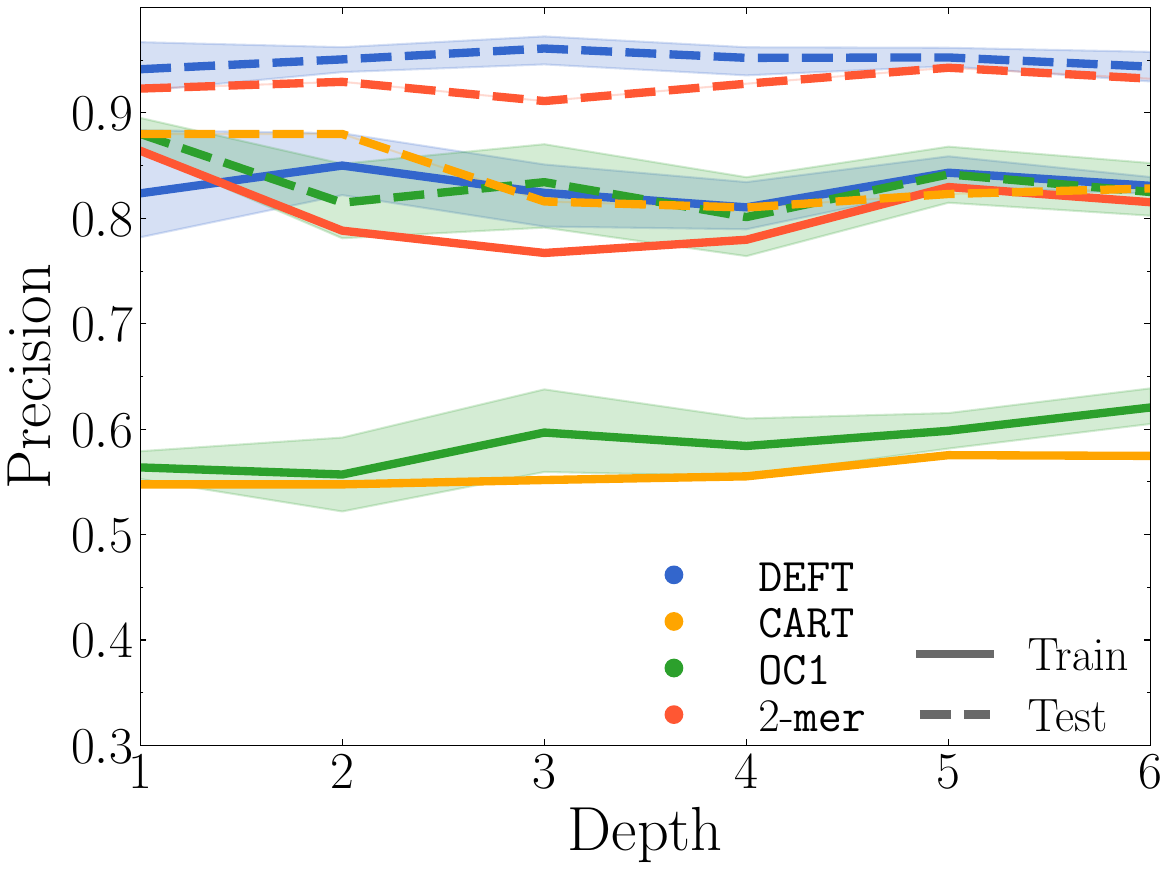}
    \caption{Precision}
    \label{fig:dsC_precision}
\end{subfigure}
\hfill
\begin{subfigure}[b]{0.24\textwidth}
    \includegraphics[width=\textwidth]{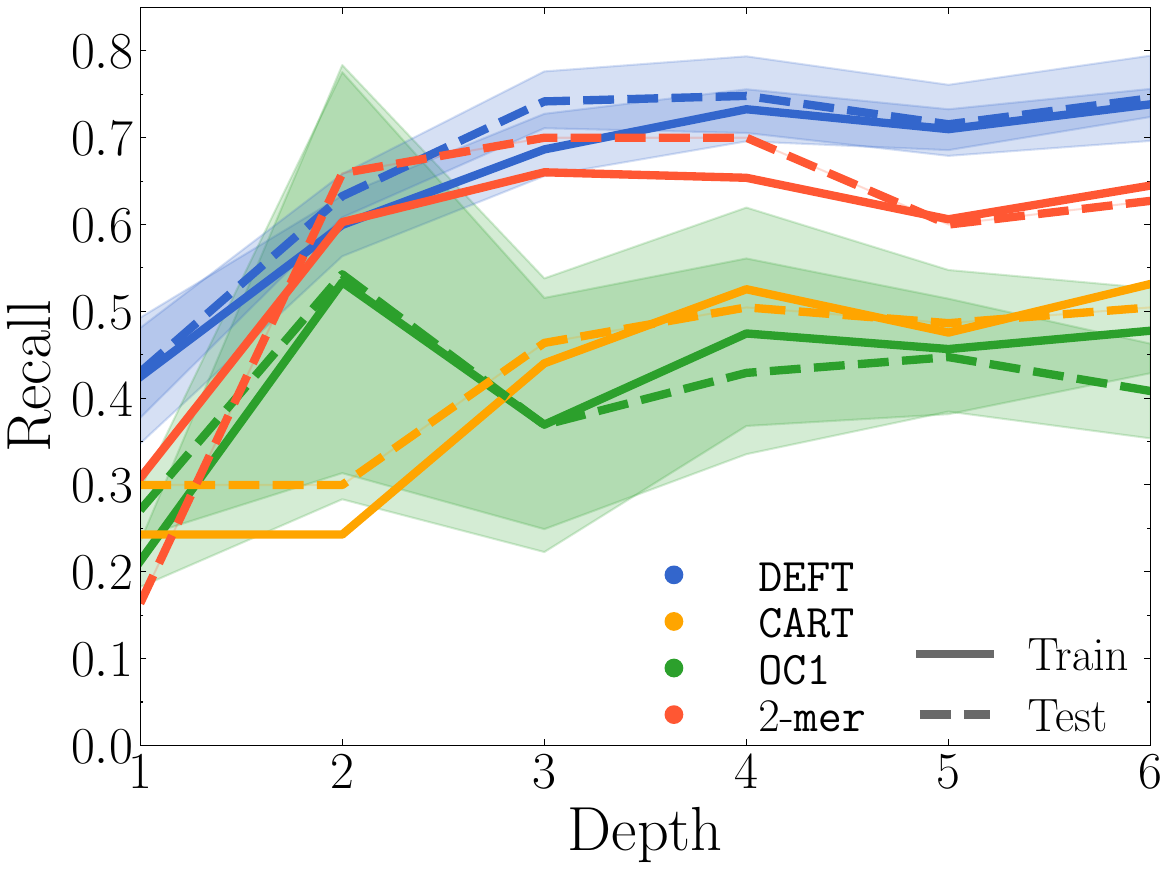}
    \caption{Recall}
    \label{fig:dsC_recall}
\end{subfigure}
\hfill
\begin{subfigure}[b]{0.24\textwidth}
    \includegraphics[width=\textwidth]{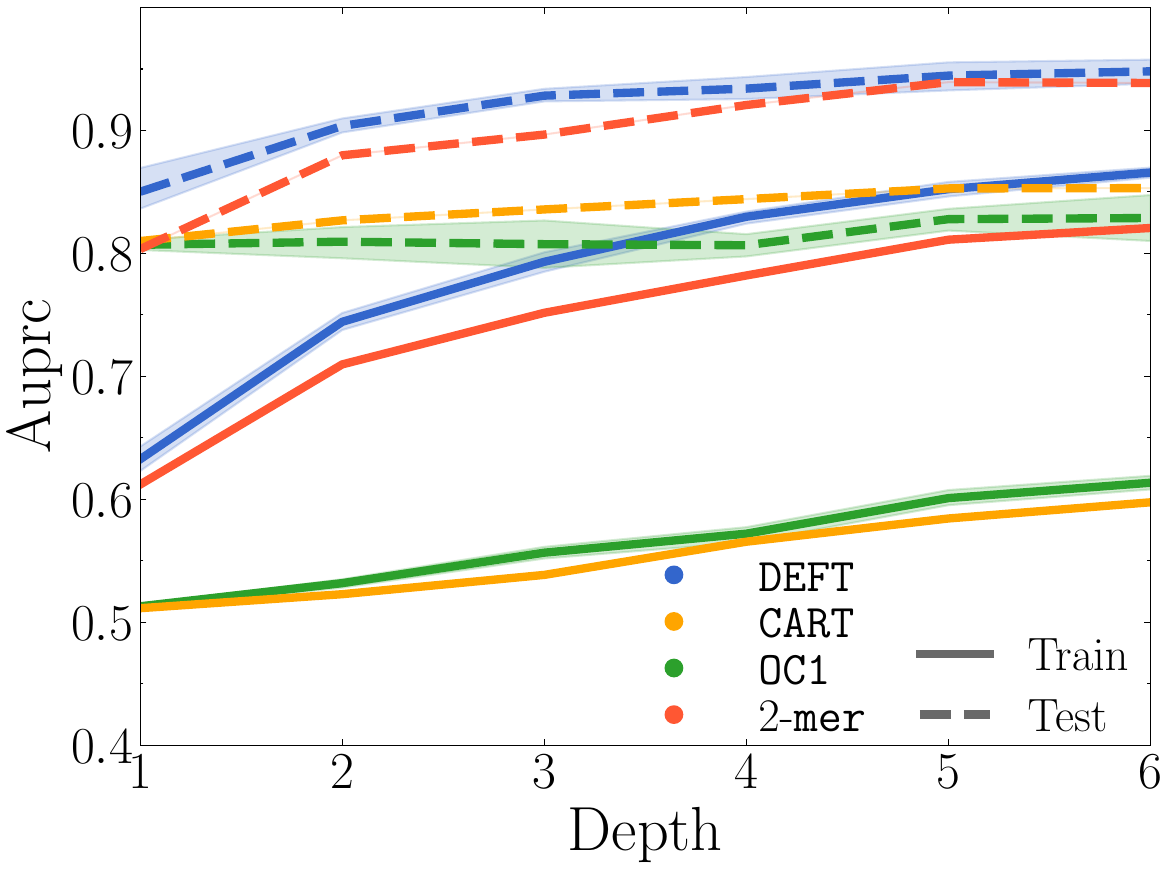}
    \caption{AUPRC}
    \label{fig:dsC_auprc}
\end{subfigure}

\caption{\textbf{Performance comparison across datasets (rows) and metrics (columns).} Each plot shows mean with $95\%$ CIs over $5$ seeds.}
\label{fig:perf_other_metrics}
\end{figure}

We report in \Cref{fig:perf_other_metrics} the \textit{F1 score}, \textit{precision}, \textit{recall} and \textit{AUPRC} for \name~ and the tree-based baselines. The results corroborate our findings from \cref{subsec:exp:performance}, demonstrating that \name~ identifies sequence features that are highly predictive both in and out of sample.

\subsection{Comparison with the Nucleotide Transformer} \label{app:nucleotide_transformer}

While the primary focus of this work is on developing \textit{transparent} models for DNA sequence analysis, we provide a comparison with a foundation model, the Nucleotide Transformer \citep{dalla2025nucleotide}, in order to contextualize \name~'s performance. It is important to note that large foundation models like the Nucleotide Transformer, while often achieving state-of-the-art predictive accuracy, operate as \textit{black boxes}, which contradicts the objective of this paper, i.e. designing models which are inherently transparent.

\textbf{Methodology.} For this comparison, we use the pre-trained \texttt{nucleotide-transformer-500m-human-ref} model \citep{nucleotide_transformer_hugging_face}, which has approximately $500$ million parameters. To adapt it for our classification tasks, we employ Low-Rank Adaptation (LoRA) \citep{hu2022lora} for fine-tuning, and we target the query and value matrices within its attention layers for LoRA adaptation. Key hyperparameters for fine-tuning are detailed in Table \ref{tab:nt_hyperparams}. The number of training steps is set to the length of the training set divided by the batch size, multiplied by the number of epochs ($2$), and we take the model which achieves the best validation AUPRC. 
The performance was then evaluated on the corresponding test sets.

\begin{table}[h!]
\centering
\caption{Nucleotide Transformer fine-tuning hyperparameters.}
\label{tab:nt_hyperparams}
\begin{tabular}{@{}ll@{}}
\toprule
Hyperparameter                   & Value                                                    \\ \midrule
LoRA rank ($r$)                    & 1                                                        \\
LoRA alpha ($\alpha$)              & 32                                                       \\
LoRA dropout                     & 0.1                                                      \\
Optimizer                        & AdamW                      \\
Learning rate                    & $5 \times 10^{-4}$                                       \\
Train batch size     & 8                                                        \\
                                        
\bottomrule
\end{tabular}
\end{table}

\textbf{Results.} The results, alongside \name~ and \texttt{CART} for reference, are presented in Table \ref{tab:nucleotide_transformer_comparison}.

\begin{table}[H]
\centering
\small
\setlength{\tabcolsep}{6pt}
\caption{\textbf{Performance comparison with the Nucleotide Transformer.}}
\label{tab:nucleotide_transformer_comparison}
\begin{tabular}{lcccccc}
\toprule
& \multicolumn{2}{c}{Pol II Pausing} & \multicolumn{2}{c}{Promoters} & \multicolumn{2}{c}{Enhancers} \\
\cmidrule(lr){2-3}\cmidrule(lr){4-5}\cmidrule(lr){6-7}
Method & Acc. & AUPRC & Acc. & AUPRC & Acc. & AUPRC \\
\midrule
\texttt{CART} ($2$-mers) & \meanstd{0.750}{0.007} & \meanstd{0.830}{0.003} & \meanstd{0.793}{0.000} & \meanstd{0.897}{0.000} & \meanstd{0.674}{0.000} & \meanstd{0.939}{0.000} \\
\name & \meanstd{0.869}{0.007} & \meanstd{0.912}{0.006} & \meanstd{0.833}{0.004} & \meanstd{0.926}{0.004} & \meanstd{0.767}{0.049} & \meanstd{0.948}{0.012} \\
\texttt{Nucleotide Transformer} & \meanstd{0.885}{0.012} & \meanstd{0.948}{0.009} & \meanstd{0.872}{0.007} & \meanstd{0.960}{0.004} & \meanstd{0.771}{0.035} & \meanstd{0.919}{0.040} \\
\bottomrule
\end{tabular}

\vspace{-2pt}
\raggedright \footnotesize
\end{table}

As anticipated, the Nucleotide Transformer, leveraging its extensive pre-training on large genomic datasets, achieves strong predictive performance. However, \name~ can significantly bridge the performance gap between \texttt{CART} and this large-scale deep learning approach. Crucially, \name~ achieves this improved accuracy while retaining the inherent transparency of decision trees and providing insights through its dynamically generated, high-level features. This suggests that \name~ offers a valuable trade-off, enhancing predictive power without sacrificing the interpretability essential for scientific discovery and hypothesis generation in DNA sequence analysis.

\subsection{Dataset-Specific Baselines} \label{app:specific_baselines}
Specialized models for each of the prediction tasks are typically variants of CNNs or attention-based networks (models that we used as baselines in our experiments in \Cref{sec:experiments}). In this section, we consider the specialized baselines \texttt{PEPMAN} \citep{feng2021machine} for the \textit{Pol II pausing} task, and \texttt{DeePromoter} \citep{oubounyt2019deepromoter} for the \textit{Promoters} task. Note that these methods are not competitors to \name since they are not interpretable by design, but we report their performance to contextualize \name.
We use the authors' implementations available on Github (\url{https://github.com/fpy94/PEPMAN} and \url{https://github.com/egochao/DeePromoter})  and we provide the results in the following tables:

\begin{table}[H]
\centering
\caption{Pol II pausing task}
\label{tab:pol2-pausing}
\begin{tabular}{lcc}
\toprule
Model & Accuracy & AUPRC \\
\midrule
\texttt{CART} (2-mers)     & \meanstd{0.750}{0.007} & \meanstd{0.830}{0.003} \\
\texttt{DEFT}     & \meanstd{0.869}{0.007} & \meanstd{0.912}{0.006} \\
\texttt{PEPMAN}   & \meanstd{0.883}{0.008} & \meanstd{0.944}{0.007} \\
\bottomrule
\end{tabular}
\end{table}

\begin{table}[H]
\centering
\caption{Promoter identification task}
\label{tab:promoter-id}
\begin{tabular}{lcc}
\toprule
Model & Accuracy & AUPRC \\
\midrule
\texttt{CART} (2-mers)         & \meanstd{0.793}{0.000} & \meanstd{0.897}{0.000} \\
\texttt{DEFT}        &\meanstd{0.833}{0.004} & \meanstd{0.926}{0.004} \\
\texttt{DeePromoter}  &  \meanstd{0.802}{0.007} & \meanstd{0.914}{0.005} \\
\bottomrule
\end{tabular}
\end{table}

As we can see, \name is competitive with respect to these deep-learning baselines, but it differs from these methods by being inherently \textit{interpretable}.

\subsection{Instantiating \name with another LLM} \label{app:gpt_oss}

\name is a modular framework, and it is not tied to a single LLM. While we used \texttt{gpt-4o} in \Cref{sec:experiments} for its strong performance, \name can be instantiated with any LLM, including stable, open-source alternatives. To demonstrate this, we reran our experiments using the open-source model \texttt{gpt-oss}. We report the test accuracy and AUPRC in the following table.

\begin{table}[H]
\centering
\small
\setlength{\tabcolsep}{6pt}
\caption{\textbf{Performance comparison with an open-source LLM.}}
\label{tab:blackbox-comparison}
\begin{tabular}{lcccccc}
\toprule
& \multicolumn{2}{c}{Pol II Pausing} & \multicolumn{2}{c}{Promoters} & \multicolumn{2}{c}{Enhancers} \\
\cmidrule(lr){2-3}\cmidrule(lr){4-5}\cmidrule(lr){6-7}
Method & Acc. & AUPRC & Acc. & AUPRC & Acc. & AUPRC \\
\midrule
\texttt{CART} ($2$-mers) & \meanstd{0.750}{0.007} & \meanstd{0.830}{0.003} & \meanstd{0.793}{0.000} & \meanstd{0.897}{0.000} & \meanstd{0.674}{0.000} & \meanstd{0.939}{0.000} \\
\name (\texttt{gpt-4o}) & \meanstd{0.869}{0.007} & \meanstd{0.912}{0.006} & \meanstd{0.833}{0.004} & \meanstd{0.926}{0.004} & \meanstd{0.767}{0.049} & \meanstd{0.948}{0.012} \\

\name (\texttt{gpt-oss}) & \meanstd{0.864}{0.009} & \meanstd{0.908}{0.006} & \meanstd{0.829}{0.001} & \meanstd{0.921}{0.002} & \meanstd{0.724}{0.031} & \meanstd{0.938}{0.010} \\
\bottomrule
\end{tabular}

\vspace{-2pt}
\raggedright \footnotesize
\end{table}

\begin{table}[htbp]
    \centering
    \caption{LLM parameters}
    \begin{tabular}{ll}
        \toprule
        \textbf{Parameter} & \textbf{Value} \\
        \midrule
        Model name &  \texttt{gpt-oss 120b} \\
        Temperature & $1$ \\
        Top p & $0.95$ \\
        Reasoning effort & Medium \\
        \bottomrule
    \end{tabular}
    \label{tab:gpt_oss_parameters}
\end{table}

As shown, \name with \texttt{gpt-oss} still significantly outperforms the \texttt{CART} baseline and remains competitive with \name instantiated with \texttt{gpt-4o}, demonstrating the framework's robustness and adaptability.

\subsection{Ensembling Trees Found by \name} \label{app:ensemble}

The randomness in the LLM's sampling process is advantageous for two reasons. It is a crucial component of the reflection mechanism, as it enables an efficient exploration across the infinite search space of possible features (cf. \Cref{app:reflection}). Furthermore, we can take advantage of this stochasticity to \textit{boost predictive performance} through ensembling. 

We show this by ensembling the predictions of three trees found by \name on the \textit{same split} for each dataset, and report the results in \Cref{tab:deft_ensemble}.

\begin{table}[H]
\centering
\small
\setlength{\tabcolsep}{6pt}
\caption{\textbf{Ensembling trees found by \name improves performance.} Results reported for a given split per dataset.}
\label{tab:deft_ensemble}
\begin{tabular}{lcccccc}
\toprule
& \multicolumn{2}{c}{Pol II Pausing} & \multicolumn{2}{c}{Promoters} & \multicolumn{2}{c}{Enhancers} \\
\cmidrule(lr){2-3}\cmidrule(lr){4-5}\cmidrule(lr){6-7}
Method & Acc. & AUPRC & Acc. & AUPRC & Acc. & AUPRC \\
\midrule
\name              & 0.870 & 0.917 & 0.830 & 0.926 & 0.713 & 0.943 \\
\name (Ensemble)  & 0.884 & 0.937 & 0.848 & 0.942 & 0.809 & 0.969 \\
\bottomrule
\end{tabular}

\vspace{-2pt}
\raggedright \footnotesize
\end{table}

We note though that this performance gain comes at the cost of the interpretability of a single tree.

\subsection{Analysis of the Reflection Mechanism} \label{app:reflection}
\textbf{Methodology.} We examine the reflection mechanism's effectiveness in exploring the complex search space of potential feature maps. For each node, we compute the mean of the scores of the features generated at each reflection iteration. We then normalize these scores across nodes to compute an averaged normalized score for each reflection iteration.

\begin{figure}[htbp]
    \centering
    \includegraphics[width=0.4\textwidth]{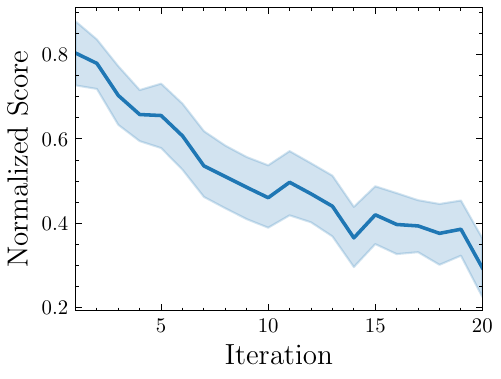}
    \caption{The reflection mechanism effectively refines the features.}
    \label{fig:reflection_scores}
\end{figure}

\textbf{Results.} \Cref{fig:reflection_scores} presents the normalized scores across reflection iterations for the Pol II pausing task, demonstrating that the reflection mechanism is essential for refining the initial candidate population. This aligns with our findings from \cref{subsec:exp:ablations}, where $\name_{\texttt{no ref}}$ exhibits lower performance compared to $\name$. This analysis also confirms that \name~ does not rely on LLM memorization to achieve good performance. Indeed, if this was the case, \name~ would be able to achieve optimal performance without the reflection mechanism, which is invalidated by \cref{subsec:exp:ablations}.

\subsection{Expert Evaluation} \label{app:expert_evaluation}
\textbf{Setup.} We extracted the top 20 features discovered by DEFT for the Pol II pausing task and asked a biologist specializing in transcriptional regulation to score each feature on a 1–5 scale along two axes:
\begin{enumerate}
    \item \textbf{Faithfulness of rationale to feature and task}: How accurately does the rationale describe what the feature’s code computes and why that computation is relevant for Pol II pausing?
    \item \textbf{Biological interpretability}: To what extent does the proposed mechanism correspond to a single, coherent biological phenomenon and appear biologically plausible?
\end{enumerate}

\textbf{Results.} The expert judged that the rationales almost always correctly described the executable feature and articulated a task-relevant mechanism, yielding a score of $4.89 \pm 0.46$ out of $5$ on faithfulness. The expert also found that \name largely proposed features plausibly aligned with determinants of pausing and transcriptional mechanics (e.g. GC-rich regions associated with pause stability or nucleosome positioning), with a score of $4.58 \pm 0.67$ out of $5$, confirming the trustworthiness of \name.

\subsection{Comparison with Foundation Models} \label{app:dnabert_hyena}
We empirically compare \name with \texttt{DNABERT-2} \citep{zhoudnabert} and \texttt{HyenaDNA} \citep{nguyen2023hyenadna} and report the results in \cref{tab:performance_comparison_dnabert}.

\begin{table}[H]
\centering
\small
\setlength{\tabcolsep}{6pt}
\caption{\textbf{Comparison with \texttt{DNABert-2} and \texttt{HyenaDNA}.}}
\label{tab:performance_comparison_dnabert}
\begin{tabular}{lcccccc}
\toprule
& \multicolumn{2}{c}{Pol II Pausing} & \multicolumn{2}{c}{Promoters} & \multicolumn{2}{c}{Enhancers} \\
\cmidrule(lr){2-3}\cmidrule(lr){4-5}\cmidrule(lr){6-7}
Method & Acc. & AUPRC & Acc. & AUPRC & Acc. & AUPRC \\
\midrule
\texttt{DNABERT-2} & \meanstd{0.795}{0.006} & \meanstd{0.881}{0.005} & \meanstd{0.937}{0.007} & \meanstd{0.967}{0.007} & \meanstd{0.855}{0.025} & \meanstd{0.979}{0.003} \\
\texttt{HyenaDNA}  & \meanstd{0.777}{0.018} & \meanstd{0.868}{0.005} & \meanstd{0.909}{0.009} & \meanstd{0.932}{0.008} & \meanstd{0.808}{0.045} & \meanstd{0.965}{0.009} \\
\texttt{CART}      & \meanstd{0.750}{0.007} & \meanstd{0.830}{0.003} & \meanstd{0.793}{0.000} & \meanstd{0.897}{0.000} & \meanstd{0.674}{0.000} & \meanstd{0.939}{0.000} \\
\texttt{DEFT}      & \meanstd{0.869}{0.007} & \meanstd{0.912}{0.006} & \meanstd{0.833}{0.004} & \meanstd{0.926}{0.004} & \meanstd{0.767}{0.049} & \meanstd{0.948}{0.012} \\
\bottomrule
\end{tabular}

\vspace{-2pt}
\raggedright \footnotesize
\end{table}

On Pol II pausing, \name outperforms both \texttt{DNABERT-2} and \texttt{HyenaDNA} while remaining fully interpretable. On Promoters and Enhancers, \texttt{DNABERT-2} and \texttt{HyenaDNA} achieve higher absolute scores, as expected for large pre-trained models, but \name substantially closes the gap between CART and these black boxes. 

\section{Broader Impacts} \label{app:broader_impacts}
This work introduces \name, an interpretable framework for DNA sequence analysis. Its positive impacts include accelerating biological research through the discovery of human-understandable sequence features, leading to greater trust in AI models applied to life sciences.
While the use of LLMs involves considerations regarding computational resources, the generation of human-interpretable features inherently facilitates expert review and validation. This opens the door for a collaborative human-AI approach, helping responsibly navigating the analysis of diverse genomic data.

\end{document}